\documentclass[lettersize,journal]{IEEEtran}
\IEEEoverridecommandlockouts                              

\usepackage{graphicx}
\usepackage{url}
\usepackage{multicol}
\usepackage{graphicx}
\usepackage{hyperref}
\usepackage[export]{adjustbox}
\usepackage{subcaption}
\usepackage{wrapfig}

\usepackage{multicol}
\usepackage{booktabs}
\usepackage{algorithm}
\usepackage[noend]{algpseudocode}
\usepackage{float}
\usepackage{amsmath, amssymb}
\usepackage{amsfonts}
\usepackage[skip=1pt,font=small,labelfont=bf]{caption}
\usepackage{bm}
\usepackage{epstopdf}
\usepackage{units}
\usepackage{physics,amsmath}

\usepackage{todonotes}
\usepackage{afterpage}

\usepackage{color}
\definecolor{purple}{RGB}{210, 0, 210}
\definecolor{international_orange}{RGB}{240, 74, 0}

\usepackage{placeins}
\usepackage{flushend}

\newcommand{\revised}[1] {\textcolor{black}{#1}} % blue
\usepackage{cite} %

\newcommand{\reals}{\mathbb{R}}
\renewcommand{\vec}{\mathbf}

\newcommand{\DeformerNet}{\emph{DeformerNet}}

\newcommand{\pcloud}{\mathcal{P}}
\newcommand{\curpcloud}{\pcloud_\mathrm{c}}
\newcommand{\goalpcloud}{\pcloud_\mathrm{g}}
\newcommand{\initpcloud}{\pcloud_\mathrm{i}}

\newcommand{\object}{\mathcal{O}}
\newcommand{\policy}{\pi}
\newcommand{\sspolicy}{\pi_\mathrm{s}}
\newcommand{\action}{\mathcal{A}}

\newcommand{\manippoint}{\vec{p}_\mathrm{m}}

\newcommand{\encodergoal}{g_g}
\newcommand{\encodercur}{g_c}

\newcommand{\featcur}{\psi_\mathrm{c}}
\newcommand{\featgoal}{\psi_\mathrm{g}}
\newcommand{\featconcat}{\psi_\mathrm{f}}

\newcommand{\deformfunc}{F}

\newcommand{\se}{\mathcal{SE}}

\title{DeformerNet: Learning Bimanual Manipulation of 3D Deformable Objects}

\author{Bao Thach\(^1\), Brian Y. Cho\(^1\), Shing-Hei Ho\(^1\), Tucker Hermans\(^{1,2}\), and Alan Kuntz\(^1\)
\thanks{$^{1}$Robotics Center and Kahlert School of Computing, University of Utah, Salt Lake City, UT 84112, USA; $^{2}$NVIDIA Corporation, Seattle, WA, USA; {\tt\{bao.thach, brian.cho, shinghei.ho, tucker.hermans, alan.kuntz\}@utah.edu}. Corresponding author: Bao Thach.}}

\begin{document}
\maketitle

\begin{abstract}
Applications in fields ranging from home care to warehouse fulfillment to surgical assistance require robots to reliably manipulate the shape of 3D deformable objects.
Analytic models of elastic, 3D deformable objects require numerous parameters to describe the potentially infinite degrees of freedom present in determining the object's shape.
Previous attempts at performing 3D shape control rely on hand-crafted features to represent the object shape and require training of object-specific control models.
We overcome these issues through the use of our novel \DeformerNet{} neural network architecture, which operates on a partial-view point cloud of the manipulated object and a point cloud of the goal shape to learn a low-dimensional representation of the object shape.
This shape embedding enables the robot to learn a visual servo controller that computes the desired robot end-effector action to iteratively deform the object toward the target shape. 
We demonstrate both in simulation and on a physical robot that \DeformerNet{} reliably generalizes to object shapes and material stiffness not seen during training,  
\revised{including \textit{ex vivo} chicken muscle tissue.}
Crucially, using \DeformerNet{}, the robot successfully accomplishes three surgical sub-tasks: retraction (moving tissue aside to access a site underneath it), tissue wrapping (a sub-task in procedures like aortic stent placements), and connecting two tubular pieces of tissue (a sub-task in anastomosis).
\end{abstract}


\begin{figure*}[t!]
    \centering
    \includegraphics[width=1.0\linewidth]{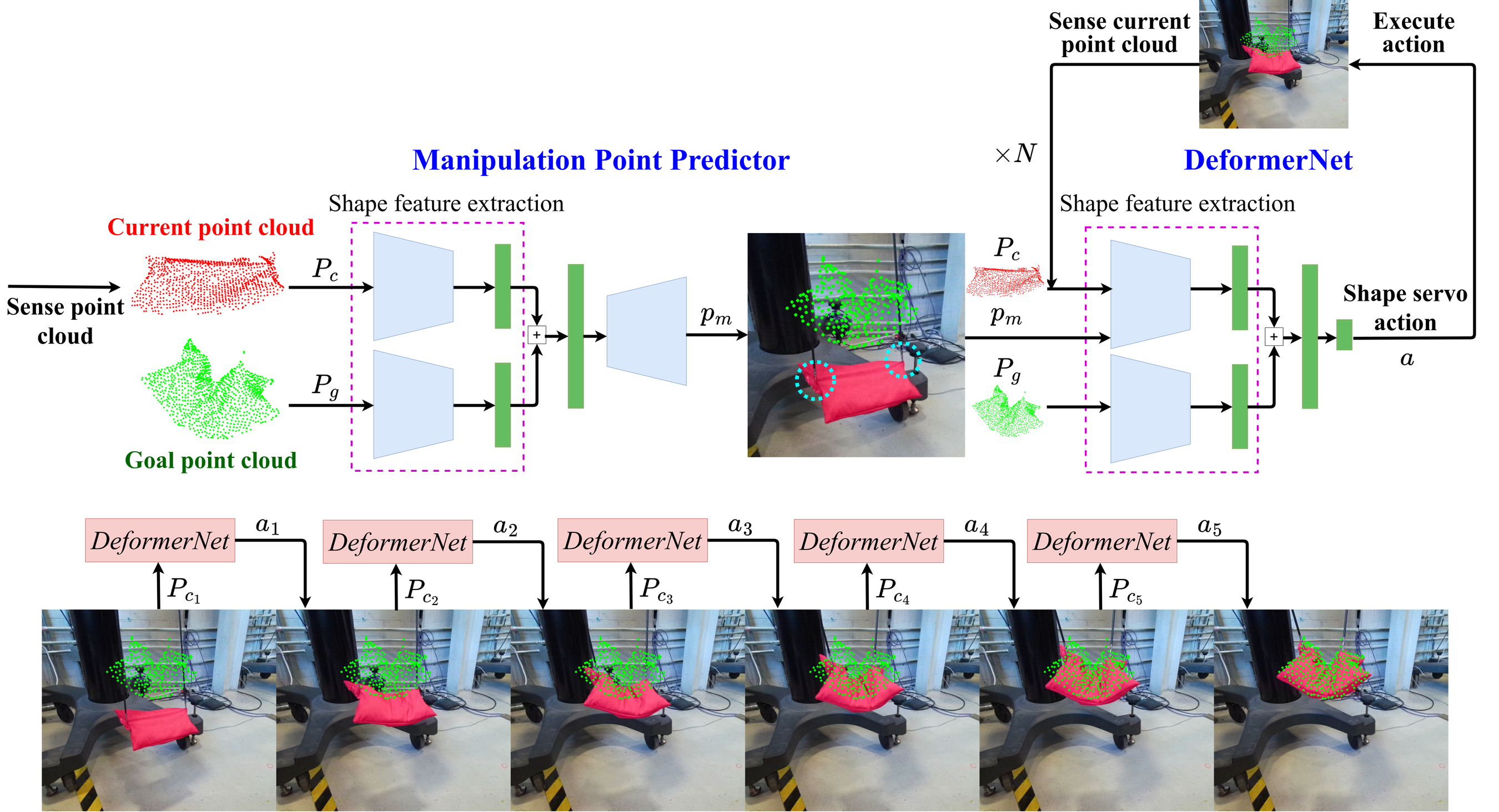}%
    \caption{\textit{(Top)} Overview of our shape-servoing-based 3D deformable object manipulation framework. Our pipeline takes as inputs the current point cloud (\(\curpcloud\)) of the deformable object as well as a goal point cloud (\(\goalpcloud\)). It then predicts where on the object the robot should grasp, i.e. manipulation point(s) $\manippoint$ (Sec~\ref{sec:mani_point}). Having grasped the object, the robot leverages our neural network $\DeformerNet{}$ to compute an action that drives the object toward the goal shape (Sec~\ref{sec:deformernet_details}). After successfully executing the action, the robot senses the current point cloud and feeds it back to $\DeformerNet{}$ to close the control loop. \textit{(Bottom)} An example manipulation sequence of a soft pillow-like object using our framework.}
    \label{fig:intro}
    \vspace{-5mm}
\end{figure*}

\section{Introduction}\label{sec:intro}

Manipulation of 3D deformable objects stands at the heart of many tasks we wish to assign to autonomous robots.
Home-assistance robots must manipulate objects such as sponges, mops, bedding, and food to help people with day-to-day life.
Robots operating in warehouses and factories must deal with many deformable materials such as bags, boxes, insulation shields, and packaging foam on a daily basis.
Surgical assistive robots must safely and precisely manipulate deformable tissue and organs. 

However, 3D deformable object manipulation presents many challenges~\cite{Sanchez2018robotic, zhu2022challenges}.
Describing shapes of deformable objects requires a potentially infinite number of degrees of freedom (DOF) compared to only 6 DOF for rigid objects.
As a result, deriving a state representation that achieves both accuracy and expressiveness is very difficult, often requiring a trade-off between the two designed by a human trial-and-error process depending on the task~\cite{zhu2022challenges}.   
Furthermore, deformable objects frequently have complex dynamics~\cite{Wu2020Learning}, making the process of deriving a model laborious and potentially computationally intensive.
These issues all present themselves in the specific problem we examine in this work: 3D deformable object shape control, i.e., tasking a robot with manipulating a 3D deformable object to reach a desired shape.

While rigid-body manipulation has received a large amount of study~\cite{mason2018toward},
autonomous 3D deformable object manipulation currently remains an under-researched area~\cite{Sanchez2018robotic,huang2021defgraspsim}---despite its potential relevance and need.
Existing work for 3D deformable shape control leverages hard-coded feature vectors to describe deformable object state~\cite{Hu20193-D}, which struggles to represent large sets of shapes.
While learning-based methods show great promise in both rigid~\cite{lu-ram2020-grasp-inference,mousavian2019graspnet} and deformable object manipulation~\cite{huang2021defgraspsim,Yan2020Learning}, these methods require a large amount of training data.
Due to the difficulty of accurately simulating deformable objects, existing methods for shape control rely on data gathered via real-world setups, limiting the efficacy of learning-based approaches.
Further, the ability to successfully manipulate deformable material is heavily dependent on where the robot grasps an object, however current approaches do not provide methods for selecting grasping points conditioned on the desired post-grasp manipulation.

In this work, we take steps toward addressing each of these gaps in the context of 3D deformable shape control.
Our method takes as input a partial-view point cloud representation of a 3D deformable object and a desired goal shape, and outputs an action that drives the object toward the goal shape (see Fig.~\ref{fig:intro} for an overview of our framework).
We build our method around a novel neural-network architecture, \DeformerNet{}, which is trained on a large amount of data gathered via a recently-developed high-fidelity deformable object simulator, Isaac Gym~\cite{Liang2018GPU,huang2021defgraspsim,macklin2019}.
Our method first reasons over the initial and target shape to select a manipulation point.  Following the selection of this grasp point, \DeformerNet{} takes the current and target point clouds of the object as well as the manipulation point location, embeds the shape into a low-dimensional latent space representation, and computes a change in end-effector pose that moves the object closer to the goal shape.
The robot executes this motion and proceeds in a closed-loop fashion generating commands from \DeformerNet{} until reaching the desired goal shape.
Figure~\ref{fig:intro} shows the initial, intermediate, and final configurations from an example manipulation using \DeformerNet{} on a physical robot.
In addition to providing the first empirical demonstration of the importance of manipulation point selection for 3D shape control,
we further develop an extended version of \DeformerNet{} for bimanual manipulation, opening the door to many applications that require more than one end-effector to accomplish tasks.

We focus our evaluation on the surgical robotics domain.
We first task a robot with manipulating three classes of object primitives into a variety of goal shapes using a laparoscopic tool.
We vary the physical dimensions and the stiffness properties of the objects.
We demonstrate effective manipulation on test objects both in simulation and on a physical robot.
We show that \DeformerNet{} outperforms both a sampling-based motion-planning strategy and a model-free reinforcement learning approach on the shape control task.

We additionally present strategies for applying our method to three common surgical sub-tasks---retraction, tissue wrapping, and tube connecting---where we derive target goal shapes from intuitive human input.
We demonstrate successful execution of these tasks both in simulation and on the physical robot. 

This work extends our prior conference paper~\cite{thach2022learning}, delivering several new contributions and a much larger number of experiments to effectively evaluate the performance of our shape servoing pipeline.
\revised{First, we develop the \textit{dense predictor}, an effective learning-based method for selecting manipulation points which we observe to perform almost as well as the ``ground-truth" manipulation points. We also compare it against a competitive alternative, the \textit{classifier}.}
Second, \DeformerNet{} takes the manipulation point location as an additional input, thus achieving substantially higher performance.
Third, \DeformerNet{} is upgraded to support changes in both robot gripper position and orientation, enabling deformable objects to reach more complex shapes. Fourth, we modify \DeformerNet{} to accommodate bimanual manipulation.
Fifth and finally, we further demonstrate the practicality of our method by leveraging \DeformerNet{} to achieve two additional surgical tasks: tissue wrapping (a sub-task in aortic stent placements), and tube connecting (a part of anastomosis).

We make available all code and data associated with this paper at \url{https://sites.google.com/view/deformernet-journal/home}.

\section{Related Work}\label{sec:related_work}
Machine learning has enhanced robots' capabilities in various challenging tasks, making it widely adopted in the robotics community.
Some existing approaches leverage machine learning with point cloud sensing to manipulate 3D rigid objects~\cite{murali2020dof,mousavian2019graspnet,deng2020self,lu-ram2020-grasp-inference,lu-iros2020-active-grasp,vandermerwe-icra2020-reconstruction-grasping}. Works propose various neural network architectures to encode object shape to achieve varying tasks such as grasp planing~\cite{mousavian2019graspnet,lu-ram2020-grasp-inference,lu-iros2020-active-grasp,vandermerwe-icra2020-reconstruction-grasping}, collision checking~\cite{murali2020dof}, shape completion~\cite{vandermerwe-icra2020-reconstruction-grasping}, and object pose estimation~\cite{deng2020self}.
In this work, we build upon these concepts to apply a learning-based approach which reasons over point cloud sensing with learned feature vectors to manipulate 3D deformable objects.

Solutions to 3D deformable object shape control~\cite{Sanchez2018robotic} can be categorized into learning-based and learning-free approaches.
Among the learning-free methods, a series of papers~\cite{Navarro-Alarcon2013b, Navarro-Alarcon2014, Navarro-Alarcon2016, qi2022model, alambeigi2018robust, alambeigi2018autonomous} define a set of geometric feature points on the object as the state representation. The authors use this representation to perform visual servoing with an adaptive linear controller that estimates the Jacobian matrix of the deformable object.
These methods, which involve precise detection of the feature points, require known objects with distinct texture and will struggle to generalize to a diverse set of objects.
Further, this formulation controls the displacements of individual points which may not fully reflect the 3D shape of the object. 
Other learning-free works~\cite{Qi2019Contour,Navarro-Alarcon2017,zhu2021vision,qi2022model} represent the object shape using 2D image contours; limiting the space of controllable 3D deformations.
Most recently, Shetab-Bushehri \emph{et al.}~\cite{shetab2022lattice} model the deformable object as a 3D lattice and successfully achieve full 3D control. However, this method requires feature correspondence between the initial and goal configuration, which may not be feasible in many real-world scenarios. 

Among learning-based 3D shape control methods, Hu \emph{et al.}~\cite{Hu20193-D} use extended Fast Point Feature Histograms (FPFH)~\cite{Rusu2010VFH} to extract a feature vector from an input point cloud and learn to predict deformation actions via a neural network to control objects to desired shapes.
However, we show in our previous work that this architecture over-simplifies the complex dynamics of 3D deformable objects and thus struggles to learn to control to a diverse set of target shapes~\cite{thach2021deformernet}. 

Among learning-based shape servoing papers for tissue manipulation specifically, Murphy \emph{et al.}~\cite{murphy2023surgical} propose an adaptive constrained optimization method to learn the Jacobian matrix of an unknown deformable tissue. However, this approach only aims to match the location of a feature point with that of a target point in image space. It is not straightforward how to scale this method to control the full 3D geometry of tissue or any arbitrary deformable object in general.
Pedram \emph{et al.}~\cite{pedram2020toward} leverage a model-free reinforcement learning approach (approximate Q-Learning). However, this method uses a set of two feature points as the state representation, which is an oversimplification of the shape servo task. The authors also did not demonstrate that their approach can generalize to different shapes of tissues. In addition, all experiments are conducted in simulation, and it is unclear how the learned policy would be transferred to a real-world manipulation scenario.

With regard to general 3D deformable object works that leverage learning-based approaches,~\cite{shen2022acid, shi2022robocraft, wi2022virdo++} employ deep neural networks to learn a dynamics model of 3D deformable objects.
There has also been work on shape control of deformable objects that are not volumetric, e.g., 1D objects, such as rope, and 2D objects, such as cloth~\cite{Wu2020Learning,matas2018sim,Yan2020Learning,ma2020contrast,McConachie2017bandit,lin2022learning}. These methods either directly learn a policy using model-free reinforcement learning (RL) that maps
RGB images of the object to robot actions~\cite{Wu2020Learning, matas2018sim} or learn predictive models of the object under robot actions~\cite{Yan2020Learning,ma2020contrast,ma2021learning,McConachie2017bandit,lin2022learning}.
These 1D and 2D works do not scale to 3D deformable objects, both because they leverage lower dimensional object and sensing (e.g. RGB images) representations as well as due to the inherent physical differences between 1D, 2D, and 3D deformable objects.

With respect to surgical robotics, several learning-based approaches have been applied to other surgical tasks including suturing~\cite{vandenberg2010_ICRA,Chiu2021Bimanual}, cutting~\cite{Thananjeyan2017_ICRA, Murali2015_ICRA}, tissue tracking~\cite{Lu2021Super,lin2022semantic},  simulation~\cite{xu2021surol}, surgical tool navigation~\cite{kim2020autonomously}, context-dependent surgical tasks~\cite{huang2021toward}, and automated surgical peg transfer~\cite{paradis2021intermittent}.
Attanasio et al.~\cite{Attanasio2020_RAL} propose the use of surgeon-derived heuristic motion primitives to move tissue flaps identified by a vision system.
In~\cite{Jansen2009_IROS}, a grasp location and planar retraction trajectory is computed with a linearized potential energy model leveraging online simulation.
In~\cite{meli2021autonomous}, a logic-based task planner is leveraged which guarantees interpretability, however the work focuses on manipulating a single thin tissue sheet and does not show shape or material property generalization or validation on a physical robot.
Nagy et al.~\cite{Nagy2018_SAMI} propose the use of stereo vison accompanied by multiple control methods, however the method assumes a thin tissue layer and a clear view of two tissue layers.
Pore et al.~\cite{pore2021safe} introduce a model-free reinforcement learning method which learns safe motions for a robot's end effector during retraction, however it does not explicitly reason over the deformation of the tissue.
We compare against a similar approach, using a model-free reinforcement learning algorithm, but adapted to our task to explicitly reason over the tissue state.
In this work, we apply our method to three surgical tasks: retraction, tissue wrapping, and tube connecting.

\section{Problem Formulation}\label{sec:problem}

We address the bimanual manipulation problem of a 3D deformable object from an initial shape to a goal shape.
In this context, \emph{3D} refers to \emph{triparametric} or \emph{volumetric} objects \cite{Sanchez2018robotic} which have no dimension significantly smaller than the other two, unlike \emph{uniparametric} (e.g., rope) and \emph{biparametric} (e.g., cloth) objects.

We define the shape of the 3D volumetric object to be manipulated as $\object \subset \reals^3$, noting that it will change over time as the robot manipulates it and the object interacts with the environment.
As typically we cannot directly sense $\object$, we consider a partial-view point cloud $\pcloud \subset \object$ as a subset of the points on the surface of $\object$, noting the prevalence of sensors that produce point clouds.
We define the point cloud representing the initial shape of the object as $\initpcloud$, the goal shape for the object as $\goalpcloud$, and the shape of the object at a given intermediate point in time $\curpcloud$.

\begin{figure}[ht]
    \centering
    \includegraphics[width=\linewidth]{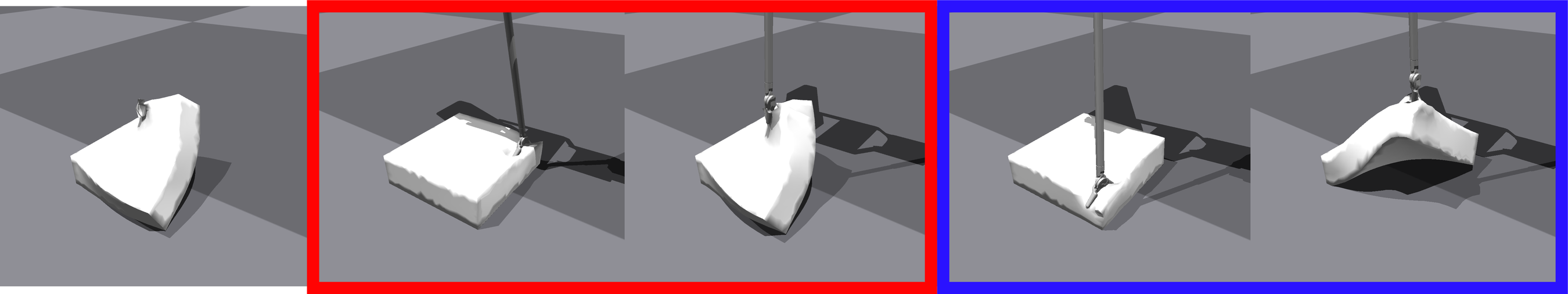} %
    \caption{Importance of manipulation point selection. Leftmost: goal shape; Red box: successful manipulation point; Blue box: failed manipulation point.}
    \vspace{-2mm}
\label{fig:mani_point}
\end{figure}

We note that the successful manipulation of a deformable object depends on the points on the object where the robot grasps, i.e., the \textit{manipulation points}. As seen from Fig.~\ref{fig:mani_point}, a poorly chosen manipulation point might make it difficult, sometimes even impossible, for the object to reach the goal shape.
Therefore, we first present the problem of selecting two manipulation points for the two manipulators, which we define as $\manippoint = [x_1, y_1, z_1, x_2, y_2, z_2] \in \object$.

Having grasped the object, the robot can change that object's shape by moving its end-effectors and in turn moving the manipulation points on the object.
We define a manipulation action $\action$ as two homogeneous transformation matrices of the two robot end-effector poses, formally $\action \in \se (3) \times \se (3)$.
The resulting problem then becomes to define a policy  $\policy: \pcloud \times \pcloud \times \manippoint \to \se (3) \times \se (3)$, which maps the point cloud representing the object shape, the goal point cloud, and the manipulation point locations, to an action describing the change in robot gripper poses that drives the object toward the goal shape, i.e., $\policy(\curpcloud, \goalpcloud, \manippoint) = \action$.
The repeated application of a successful policy $\policy$ results in a manipulation trajectory which, when executed by the robot, results in transforming the object from its initial shape to a goal shape.

The problem can be simply reduced to the single manipulator case if required by redefining $\manippoint = [x_1, y_1, z_1] \in \object$ and $\action \in \se (3)$.

\section{Learning-Based Shape Servoing}\label{sec:method}

The shape servo formulation~\cite{Navarro-Alarcon2017,Hu20193-D} uses sensor feedback, here in the form of partial-view point clouds of the manipulated object, as input to a policy that computes a robot action that attempts to deform the current shape, \(\curpcloud\) closer to the target shape, \(\goalpcloud\).

Building upon the above general formulation, we develop our novel learning-based shape servoing framework (see Fig.~\ref{fig:intro} for a visual overview). First, from point cloud observations of \(\curpcloud\) and \(\goalpcloud\), we leverage a neural network to select good manipulation points for both manipulators (Sec.~\ref{sec:mani_point}).
Second, our neural network $\DeformerNet{}$ (Sec.~\ref{sec:deformernet_details}) computes the desired robot action, using \(\curpcloud\), \(\goalpcloud\), and the manipulation points as inputs.
We perform shape servoing via the repeated application of $\DeformerNet{}$, taking an action, sensing the new state, determining a new action, etc., until convergence.

\subsection{\DeformerNet{} Architecture Details}\label{sec:deformernet_details}

\begin{figure*}[ht]
    \centering
    \includegraphics[width=0.95\linewidth]{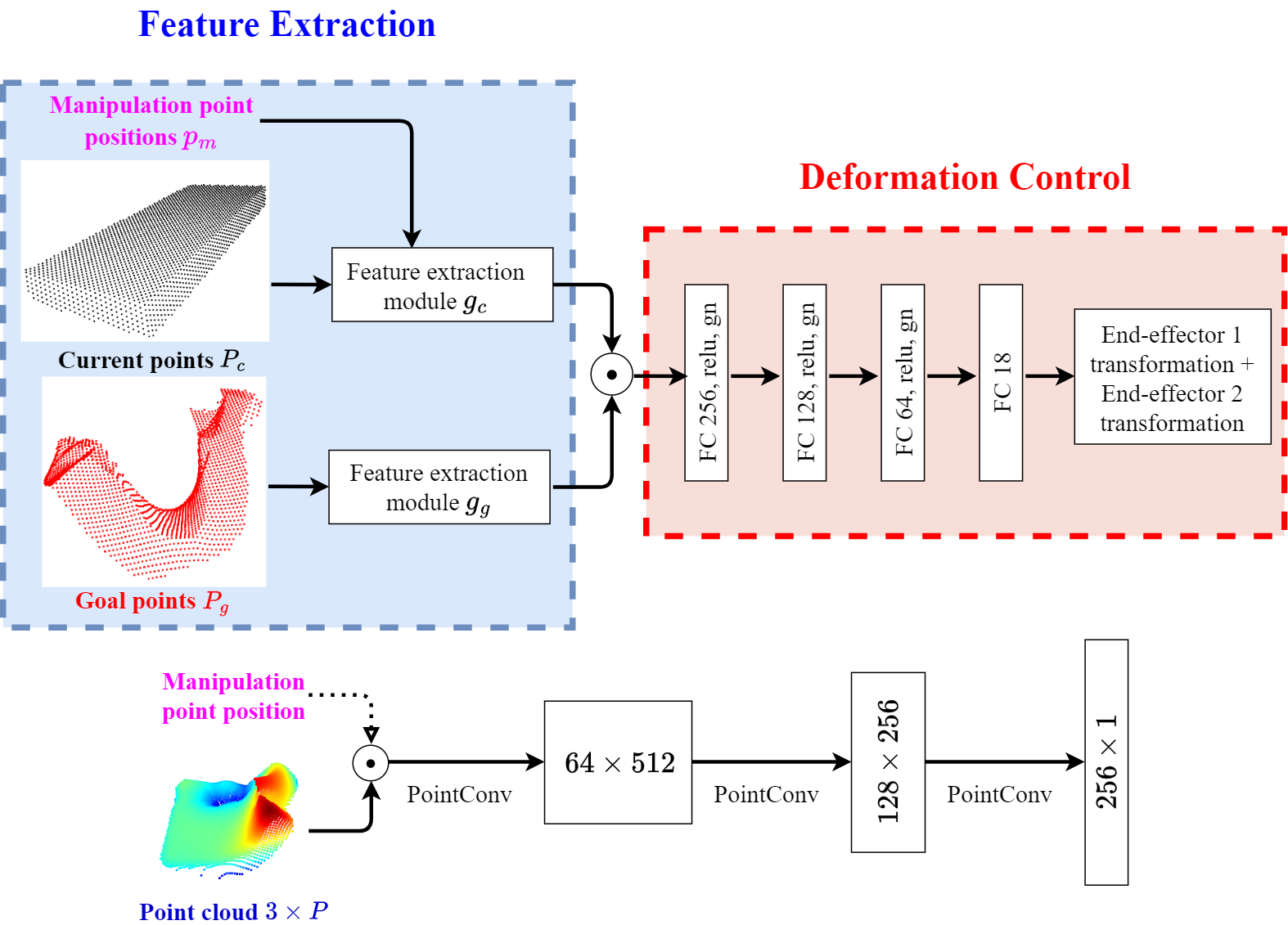}%
    \caption{\textbf{(Top)} Architecture of \DeformerNet{}. Bounded by the dotted blue box is the \textit{feature extraction} stage, and bounded by the dotted red box is the \textit{deformation control} stage. The \textit{feature extraction} stage takes as inputs the current point cloud (\(\curpcloud\)) as well as the goal point cloud (\(\goalpcloud\)), passes each of them through its corresponding feature extraction module, and eventually produces two 256-dimensional vectors. These vectors are concatenated together to compose a final 512-dimensional \textit{shape feature vector}. The \textit{deformation control} stage takes in this shape feature vector, passes it through a series of fully-connected layers, and finally outputs an action that drives the object toward the goal shape.    
    \textbf{(Bottom)} Architecture of the feature extraction module. It consists of three sequential PointConv~\cite{PointConv2019} convolutional layers. The feature extractor $\encodergoal$ of \(\goalpcloud\) only takes the point cloud as input. For the current point cloud \(\curpcloud\), $\encodercur$ takes in additionally the two manipulation point positions.} 

    \label{fig:bimanual_DeformerNet}

\end{figure*}

$\DeformerNet{}$ performs as a shape servo policy of the form \(\sspolicy(\curpcloud, \goalpcloud, \manippoint) = \action \). We decompose our policy into two stages: (1) a feature extraction stage and (2) a deformation controller (see Fig.~\ref{fig:bimanual_DeformerNet} top).

We use two parallel feature extraction channels that take as inputs \(\curpcloud\) and \(\goalpcloud\) and generate two feature vectors $\featcur = \encodercur(\curpcloud, \manippoint)$ and $\featgoal = \encodergoal(\goalpcloud)$ respectively. The feature extractor $\encodergoal$ of \(\goalpcloud\) only takes the point cloud as input. The feature extractor $\encodercur$ of \(\curpcloud\) takes as inputs both the point cloud and two vectors encoding the two manipulation point locations. Details about how to obtain these manipulation point vectors will be presented later in this section. We then concatenate the two feature vectors to obtain the final feature vector: \(\featconcat = \featcur\bigodot\featgoal\). 
Our deformation control function, \(\deformfunc\), takes this feature vector as input and outputs the desired change in end-effector poses, hence:
\(\action = \deformfunc(\featconcat)\).

The composite shape servo policy thus takes the form \(\sspolicy(\curpcloud, \goalpcloud, \manippoint) =  
\deformfunc(\encodercur(\curpcloud, \manippoint)\bigodot\encodergoal(\goalpcloud)) = \action\).
\revised{Executing actions output by our shape servo policy \(\sspolicy\) involves a two-level robot controller architecture. At the high level, the desired end-effector transformations generated by \DeformerNet{} are fed into a resolved-rate controller to calculate a trajectory of desired joint velocities for the robot. Successful execution of this trajectory will bring the robot end-effectors to the target poses.
At the lowest level, the joint-level controller is responsible for controlling each joint to achieve the desired velocities determined above.
We execute our learning-based policy in a closed-loop manner, as depicted in Fig~\ref{fig:intro}. Our process begins by sensing the initial point cloud of the object and feeding it into \DeformerNet{} to derive the first desired action. As a result of executing this action, the object transitions to a new deformed state. We then sense a new, current object point cloud and once again leverage \DeformerNet{} to compute a new action. If this new action magnitude surpasses a defined threshold $\epsilon$, we execute it, otherwise we terminate the operation (leveraging action magnitudes smaller than $\epsilon$ as a metric of convergence). This cycle persists as we continue to sense new point clouds and execute the associated actions, as long as the action magnitude is above the threshold.}

Training \DeformerNet{} takes a straightforward supervised approach. \revised{We simply record the robot manipulating objects in simulation with diverse geometries and stiffnesses to enable generalization to the variety of objects the robot will need to manipulate in deployment. The specific manipulation strategy can be for example, random.} We then set the terminal object point cloud as \(\goalpcloud\), select any previous point cloud from the trajectory as \(\curpcloud\) and the associated end-effector transformation between the two configurations as \(\action\).
We give further details of this training procedure in Sec.~\ref{sec:experiments}.

Before discussing details of the \DeformerNet{} architecture, let us illustrate the formal definition of a point cloud. A point cloud of dimension $c\times n$ is a set of $n$ points in which each point has $c$ features. These features could be derived directly from a sensor (such as 3D position $(x,y,z)$, surface normal, and color), or learned from a neural network.

\revised{We adopt an object-centric coordinate frame for all point clouds, with its origin located at the centroid of the object in its undeformed state. In all simulation and physical robot experiments, we first place the undeformed object in the scene and then fit a bounding box around the object's partial point cloud. This bounding box is generated using the Trimesh library~\cite{trimesh2019}. Subsequently, we assign the bounding box center as the origin of the object-centric frame. We define the y-axis as the principal axis of the object point cloud, as indicated by the bounding box. The z-axis is oriented as the vector opposite to gravity, and the x-axis is computed via the cross-product of the y and z axes.}

Fig.~\ref{fig:bimanual_DeformerNet} visualizes the full architecture of \DeformerNet{}. 
Prior to training or running \DeformerNet{}, we first downsample the input point cloud, \(\curpcloud\), and goal point cloud, \(\goalpcloud\), to 1024 points using the furthest point sampling method~\cite{Qi2017PointNet}. These two point clouds, each with shape $3\times1024$, are then fed to the neural network as input. We also input the manipulation point locations to \DeformerNet{} in the form of two manipulation point channels of shape $1\times1024$ concatenated with the current point cloud (shape $3\times1024$) to create a cloud of shape $5\times1024$. These two channels are encoded by giving a value of $1$ to the $50$ points on the current point cloud nearest to the manipulation point, and a value of $0$ to the other points.
Each feature extractor uses three sequential PointConv~\cite{PointConv2019} convolutional layers that successively output point clouds of dimension $64\times512$, $128\times256$
and ultimately a 256-dimensional vector that acts as the shape feature.
We concatenate the shape features of the current and goal point cloud together to form a final 512-dimensional \textit{shape feature vector}.

The deformation control stage takes the 512-dimension \textit{shape feature vector} and passes it through a series of fully-connected layers (256, 128, and 64 neural units, respectively). The fully-connected output layer produces an 18-dimensional vector. The first six dimensions account for the desired position displacements of the two grippers. The remaining twelve-dimensional vector is split into two 6D vectors, each mapped back to a $3\times3$ rotation matrix, using the method in~\cite{zhou2019continuity}---\cite{zhou2019continuity} conducted extensive studies where they showcased that this 6D representation is better for learning rotation than other alternatives such as quaternions, axis-angles, or Euler angles. These matrices represent the transformation between the current grippers' orientations and the desired orientations. Together the 18-dimensional output vector constructs the homogeneous transformation matrices between the current poses and the desired poses of the two robot end-effectors. We use the ReLU activation function and group normalization~\cite{wu2018group} for all convolutional and fully-connected layers except for the linear output layer.

We further note that by simply modifying the input and output of \DeformerNet{}, we can achieve the single-arm manipulation of deformable objects. This version of \DeformerNet{} takes as inputs the current and goal point cloud as well as the manipulation point location of the robot. It outputs a 9-dimensional vector, which can be converted into the homogeneous transformation matrix between the current end-effector pose and the desired pose as above. 
Broadly speaking, in theory \DeformerNet{} can generalize to any number of manipulators.

\subsection{Manipulation point prediction details}\label{sec:mani_point}

\revised{As discussed above and shown in Fig.~\ref{fig:mani_point}, the location at which the robot grasps the object greatly influences whether the robot will be able to manipulate a deformable object to the target shape. As such we present here the \textit{dense predictor}, a learning-based approach to effectively selecting an appropriate manipulation point prior to performing the shape servoing task. 
This is a popular concept commonly used for generating grasp poses for rigid objects~\cite{asif2019densely, cheng2020high}. Here we build on this approach to select manipulation points for deformable objects.} 

\begin{figure}[ht]
    \centering
    \includegraphics[width=1.0\linewidth]{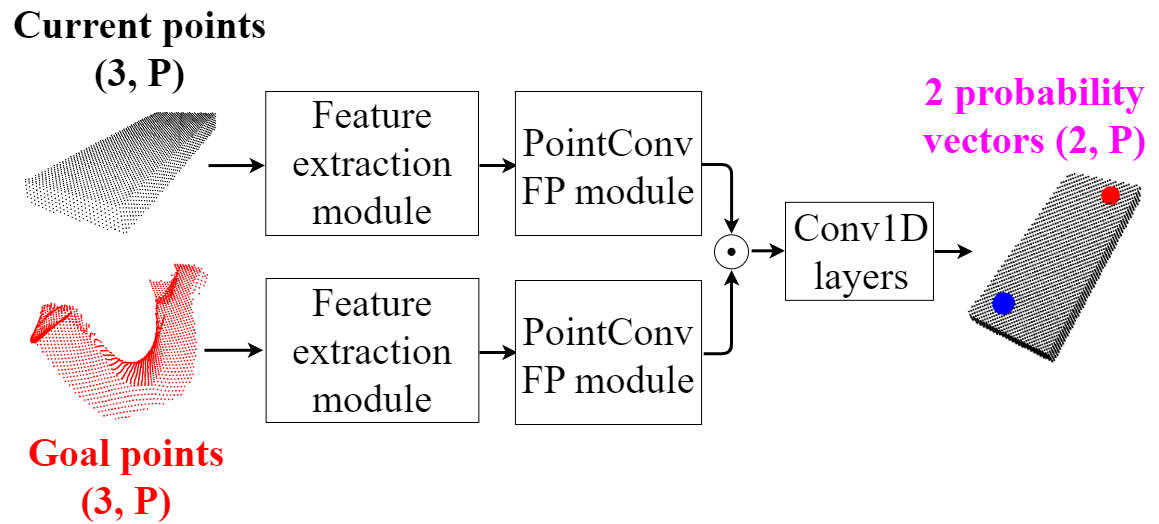}%
    \caption{Architecture of the \textit{dense predictor} network, \revised{our manipulation point selection method}. We use the feature extraction module of $\DeformerNet$ as the encoder, and the \textit{feature propagation} (FP) module of PointConv~\cite{PointConv2019} as the decoder. The outputs of the encoder-decoder series are concatenated together to form a feature point cloud of shape $128\times P$, then passed through a series of 1D Convolution layers to output two vectors of shape $1\times P$, and finally normalized to 0 to 1. The two manipulation points are then defined as those with the highest value in each vector.   
    The red and blue spheres represent where the \textit{dense predictor} predicts to be the best manipulation points for the two robot arms.}
    \label{fig:dense_predictor}
\end{figure}

Recall that we wish to find two manipulation points for two robots on the surface of the object, \(\manippoint \in \object\). However, we must infer this location given the initial \(\initpcloud\) and target point cloud \(\goalpcloud\), prior to acting. 
\revised{In our \textit{dense predictor} method}, we leverage a neural network with an encoder-decoder architecture to learn the manipulation point. The architecture of this network is visualized in Fig.~\ref{fig:dense_predictor}. The encoder is the same as the feature extraction module of $\DeformerNet$. For the decoder, we utilize the \textit{feature propagation} module (FP module) of PointConv~\cite{PointConv2019}. The neural network takes as input the current and goal point cloud. It then passes each point cloud through the encoder-decoder series and eventually outputs two point clouds of shape $64\times P$, where $P$ is the number of points in the current and goal point cloud. These two point clouds are then concatenated together to form a feature point cloud of shape $128\times P$. Finally, we pass it through a series of 1D Convolution layers to output a point cloud of shape $2\times P$, which is equivalent to two vectors, each with size equal to the number of points in the current point cloud. These vectors are separately normalized to 0 to 1 using the softmax function. 
Each vector contains $P$ values, representing the likelihood of every point in the current point cloud being a good manipulation point.
The two manipulation points can then be straightforwardly defined as the two points with the highest likelihoods in each vector.

\revised{A competitive learning-based alternative to the \textit{dense predictor} is the \textit{classifier}, which is also a popular technique for generating grasp poses for rigid objects~\cite{mousavian2019graspnet,lu-ram2020-grasp-inference,lu-iros2020-active-grasp, vandermerwe-icra2020-reconstruction-grasping}. We adapt this technique to deformable object manipulation, and conduct a comparative study between the \textit{dense predictor} and the \textit{classifier} in Sec.~\ref{sec:mani_point_selection}.}

\section{Goal-Oriented Shape Servoing Experiments}\label{sec:experiments}

\begin{figure}[ht!]
    \centering
    \includegraphics[width=1\linewidth]{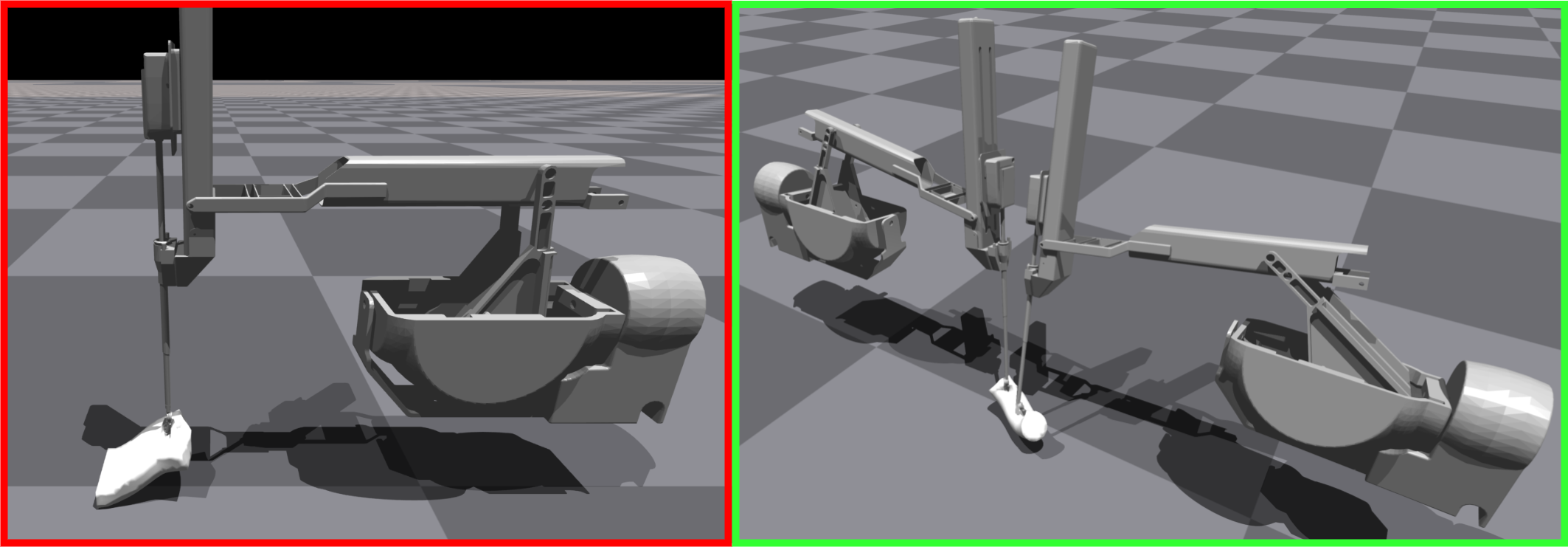} 
    \includegraphics[width=0.49\linewidth]{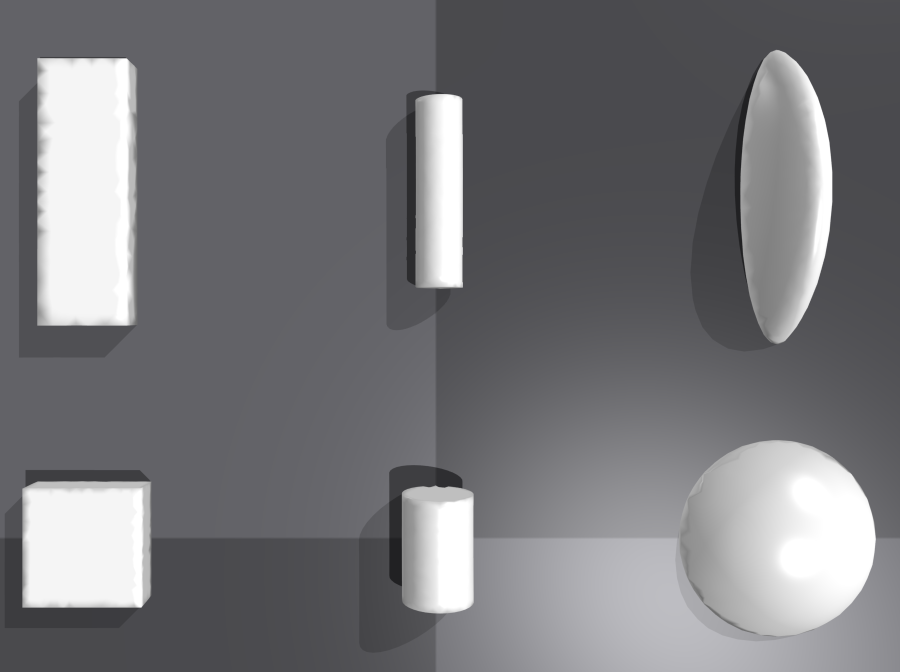}  
    \includegraphics[width=0.45\linewidth]{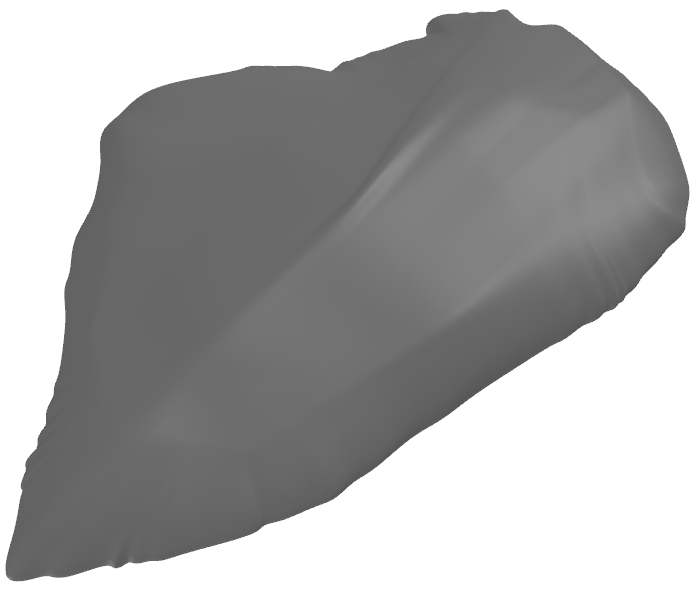}  
    
    \caption{(Top left) Simulation setup for single-arm \DeformerNet{} experiments, showing a patient-side manipulator of the dVRK in Issac gym. (Top right) Simulation setup in Issac gym for dual-arm \DeformerNet{} experiments.
    (Bottom left) We train and test on a diverse set of object geometries. Here we provide some sample objects from the training dataset. Leftmost are the two box objects with an aspect ratio of 1 and 3, respectively. In the middle are the two cylinder objects with an aspect ratio of 3 and 8, respectively. Rightmost are the two hemi-ellipsoid objects with an aspect ratio of 1 and 4, respectively. \revised{(Bottom right) We also challenge our method with manipulating chicken muscle tissue, an object with complex geometry that was \textit{unseen} during training and outside of the training set distribution.}}
    \label{fig:display_objects}
\end{figure}

\begin{figure}[ht!]
    \centering
         \includegraphics[width=0.9\linewidth]{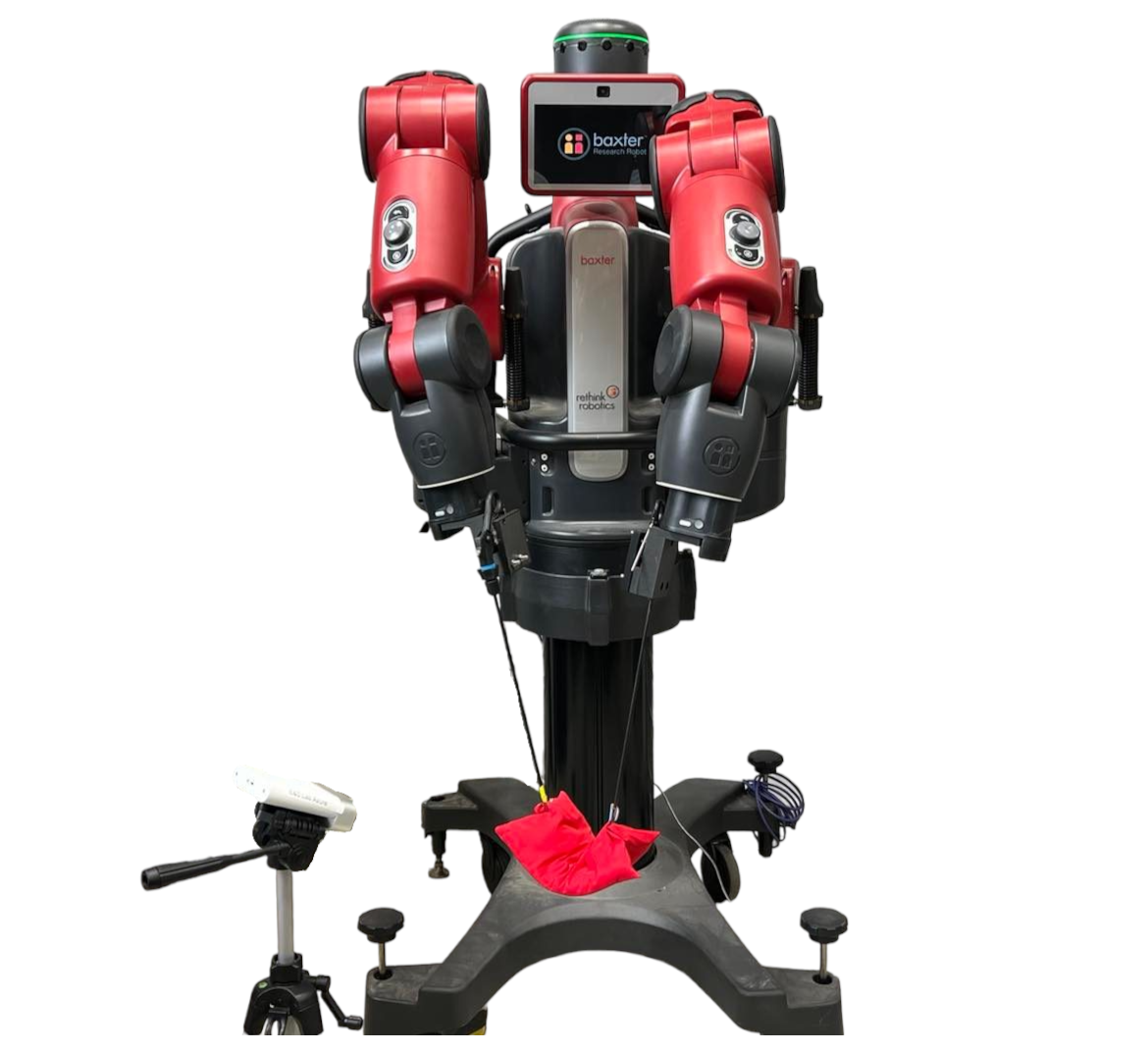}%
    \caption{Physical robot experiment setup for the shape servoing task. The setup includes a bimanual robotic system, an RGBD camera, and a deformable object.}
    \label{fig:exp_setup}
\end{figure}

We evaluate our method in both simulation, via the Isaac Gym environment \cite{Liang2018GPU}, and on a real robot.
For both simulation and real robot experiments, training data for the learned models are generated in Isaac Gym.
In Isaac Gym, we use a simulation of a patient-side manipulator of the daVinci research kit (dVRK)~\cite{Kazanzides2014_ICRA_DVRK} robot to manipulate objects (see Fig.~\ref{fig:display_objects}).
For the real robot experiments, we use a Baxter research robot with a laparoscopic tool attached to its end effector and an Azure Kinect camera \revised{to gather} point clouds of the deformable object (see Fig.~\ref{fig:exp_setup}).
In this section, we will first examine the performance of single-arm \DeformerNet{}, before moving on to conduct an evaluation of the more complex dual-arm setting.

\subsection{Goal-Oriented Shape Servoing in Simulation}
We first evaluate our method's ability to deform objects to goal point clouds in simulation.
In our previous workshop paper~\cite{thach2021deformernet}, we reported the performance of our method when the model was trained and tested on one object geometry with constant stiffness (Young's modulus) and demonstrated that our method outperforms a current state-of-the-art method for learning-based 3D shape servoing~\cite{Hu20193-D}.

We expand on this evaluation in this work by first evaluating our method's ability to control the shape of a variety of 3D deformable object shape primitives.
We evaluate three primitive shape types, rectangular boxes, cylinders, and hemi-ellipsoids (see Fig.~\ref{fig:display_objects}, bottom).
For each primitive shape type, we investigate three different stiffness ranges (characterized by their Young's modulus): $\unit[1]{kPa}$, $\unit[5]{kPa}$, and $\unit[10]{kPa}$, which represent stiffness properties similar to those seen across different biological tissues~\cite{hinz2012mechanical,handorf2015tissue}. More details about these stiffness ranges will be provided in Sec.~\ref{sec:training_data_generation}

\subsubsection{Neural Networks' Training Details}\label{sec:training_data_generation}

With each of the nine object categories, we create a training dataset of multiple objects with diverse geometries \revised{and stiffnesses} by performing the following procedures. For the box primitive, we uniformly sample a width value, a thickness value, and an aspect ratio, then multiply the width and the aspect ratio together to get the height. For the cylinder primitive, we uniformly sample a radius value and an aspect ratio, then multiply them together to get the height. For the hemi-ellipsoid primitive, we first uniformly sample a radius value to create a hemisphere; we then sample an aspect ratio, which is used to narrow one axis to create a more interesting hemi-ellipsoid geometry. We visualize example objects from the training dataset in Fig.~\ref{fig:display_objects} (bottom).
In addition, each object for training is assigned a Young's modulus sampled from a Gaussian distribution of $\mathcal{N}(\unit[1]{kPa}, \unit[0.2^2]{kPa})$, $\mathcal{N}(\unit[5]{kPa}, \unit[1^2]{kPa})$, or $\mathcal{N}(\unit[10]{kPa}, \unit[1^2]{kPa})$ for the $\unit[1]{kPa}$, $\unit[5]{kPa}$, and $\unit[10]{kPa}$ test scenarios, respectively. We train a separate model for each of the nine object types, (3 geometric types $\times$ 3 stiffness ranges) using the same \DeformerNet{} architecture.

To generate each training dataset for the bimanual manipulation setup, we first randomly sample 300 initial object configurations (geometry and stiffness). Then with each configuration, we obtain two random manipulation points by sampling two points on the object's surface and grasping the object there. 
For each pair of initial configuration and initial manipulation points, the robot deforms the object to 10 random shapes by moving its end-effectors to 10 random poses, for a total of 3000 random trajectories.
We record 1) partial-view point clouds of the object and 2) the robot’s end-effector poses, at multiple checkpoints during the execution of this trajectory using the simulated depth camera available inside the Issac gym environment. Specifically, we pause the simulation every 15 simulation frames and record new observations.
To generate training datasets for the single-arm manipulation case, we follow almost the exact same procedure except for only using one end-effector to deform the object to random shapes.

We now explain how to leverage this data to form supervised input-output pairs for training \DeformerNet{}. 
A trajectory with $M$ recorded checkpoints would include $M$ recorded point clouds $\pcloud_1,\ldots,\pcloud_M$; $M$ recorded manipulation point ${\manippoint}_1,\ldots,{\manippoint}_M$; and $M$ recorded end-effector poses $\vectorbold{x}_1,\dotsc,\vectorbold{x}_M$.
The input to $\DeformerNet$ consists of a point cloud along the trajectory $\pcloud_i$ (initial shape), the point cloud at the end of this trajectory $\pcloud_M$ (goal shape), and the selected manipulation point ${\manippoint}_i$, for $i=1,\dotsc,M$. 
The output of $\DeformerNet$ is computed as the homogeneous transformation matrix between the corresponding end-effector pose $\vectorbold{x}_i$ and that at the end of trajectory $\vectorbold{x}_M$.
We sample 20,000 such pairs of data points for training $\DeformerNet$. 

To learn the \textit{dense predictor}, we leverage the same data, slightly modified. The input becomes \((\pcloud_i, \pcloud_M)\), and the output is the selected manipulation point ${\manippoint}_i$. We use 20,000 data points for training the \textit{dense predictor}.

Training the \textit{classifier} requires both examples of successful (positive samples) and failed manipulation points (negative   samples). 
To obtain the training data for the \textit{classifier}, we start with the same dataset for training $\DeformerNet$ and then augment it using the following procedure to derive the positive and negative samples.
First, from the 1024 points of the downsampled \(\pcloud_i\), we sort them based on their distances to the ground-truth manipulation point ${\manippoint}_i$. Second, we sample 5 points from the 50 nearest points to ${\manippoint}_i$ and define them as successful manipulation points: ${\manippoint}_{i+_{j}}$, for $j=1,\dotsc,5$. Third, we sample 5 points from the 800 furthest points and set them to be negative samples: ${\manippoint}_{i-_{j}}$, for $j=1,\dotsc,5$. 
As a result, for every data point from the original dataset, we can derive 10 data points for the \textit{classifier}. Finally, we form supervised input-output pairs. For the positive data points, the input is \((\pcloud_i, \pcloud_M, {\manippoint}_{i+_{j}})\) and the output is $1$. For the negative data points, the input is \((\pcloud_i, \pcloud_M, {\manippoint}_{i-_{j}})\) and the output is $0$. We sample 100,000 data points for training the \textit{classifier}.

From our experimental results (which will be presented formally in detail later in this paper), we observe that the \textit{dense predictor} is a superior method to the \textit{classifier}. Even though they have comparable performance, running \textit{dense predictor} is faster due to the nature of its neural network architecture. Therefore in this paper, we choose the \textit{dense predictor} to be our primary manipulation point selection method.

To train \DeformerNet{}, we use the standard mean squared error loss function for the position component of the 18-dimensional output vector (first 6 elements), and geodesic loss for the orientation component (last 12 elements). 
To train the \textit{dense predictor} and the \textit{classifier}, we use the standard cross-entropy loss. 

For all \DeformerNet{}, \textit{dense predictor}, and \textit{classifier}, we train the neural networks end-to-end. We adopt the Adam optimizer~\cite{kingma2014adam} and a decaying learning rate which starts at $10^{-3}$ and decreases by 1/10 every 50 epochs.

\subsubsection{Evaluation Metrics}\label{sec:eval_metrics}
We use two distance metrics to evaluate how close a final object shape (after running our shape servoing framework) is to the goal shape.
These two distance metrics are Chamfer distance and node distance.
Chamfer distance computes the difference between the final object point cloud and the goal point cloud by summing the distances between each point in one point cloud and its closest point in the other point cloud:
\begin{equation}
d(\mathcal{P}_f, \mathcal{P}_g) = \sum_{x\in \mathcal{P}_f}\min_{y\in \mathcal{P}_g} ||x - y||^2 + \sum_{y\in \mathcal{P}_g}\min_{x\in \mathcal{P}_f} ||x - y||^2
\end{equation}
Node distance measures the difference between the final shape and goal shape by averaging the Euclidean distances between each pair of corresponding ``particles":
\begin{equation}
d(\mathcal{P}_f, \mathcal{P}_g) = {1\over |\mathcal{P}_g|}\sum_{x_f \in \mathcal{P}_f, x_g \in \mathcal{P}_g} ||x_f - x_g||^2, 
\end{equation}
where $x_f$ and $x_g$ are corresponding ``particles" that belong to the final object shape and goal object shape respectively.
Node distance can be computed in simulation due to the fact that Issac Gym represents each deformable object as a set of particles located on the object surface, which can be interpreted as the object's full point cloud. The indices of these particles are also fixed throughout simulations, thus giving us access to correspondences. 
Node distance is a much more reliable metric than Chamfer distance because it has access to both the object full geometry and the correspondence between particles in the current shape and those in the goal shape, whereas Chamfer distance calculation can only leverage partial-view point clouds and an estimate of correspondence (using the nearest neighbor).
However, node distance is only available in simulation, because for the physical robot experiments we do not have access to the particle information or correspondence.

\revised{We additionally use the total number of steps as an evaluation metric to assess the performance of our shape servoing pipeline. We define this metric as the number of actions queried from \DeformerNet{} before convergence. For any manipulation sequence, we want this value to be as low as possible. Number of steps equal to 1 means that it only takes the robot a single action to complete the shape servoing process.}

\subsubsection{Performance on Novel Shape Servoing Test Scenarios}\label{sec:generalization_performance} 
We will first examine the performance of single-arm \DeformerNet{}. For each of the nine object categories, we evaluate the performance of our method on 100 novel goal shapes, which are generated using the following procedure. First, we sample 10 new objects \revised{unseen during training, each} with a different geometry and stiffness. Then, for each object, we command the robot to manipulate the object into 10 random shapes and record these as the test goal point clouds. 
\revised{We emphasize that, after generating the random goal shapes, we only feed the recorded goal point cloud to \DeformerNet{}. The actual actions that achieved those are discarded and not known by the method. The object is manipulated to the goal via a system with no knowledge of how that goal was generated.}
\revised{Additionally, it is important to emphasize that, although all test objects and test goal shapes are sampled from the same distributions as the training dataset, the specific object geometries, stiffness values, and goal shapes tested on are entirely \textit{unseen} during training.}

We run our shape servoing framework on the above  $100\times9=900$ test goal shapes. For all cases, we select the manipulation point using our dense predictor method.

\begin{figure}[ht]
    \centering

    \includegraphics[width=1\linewidth]{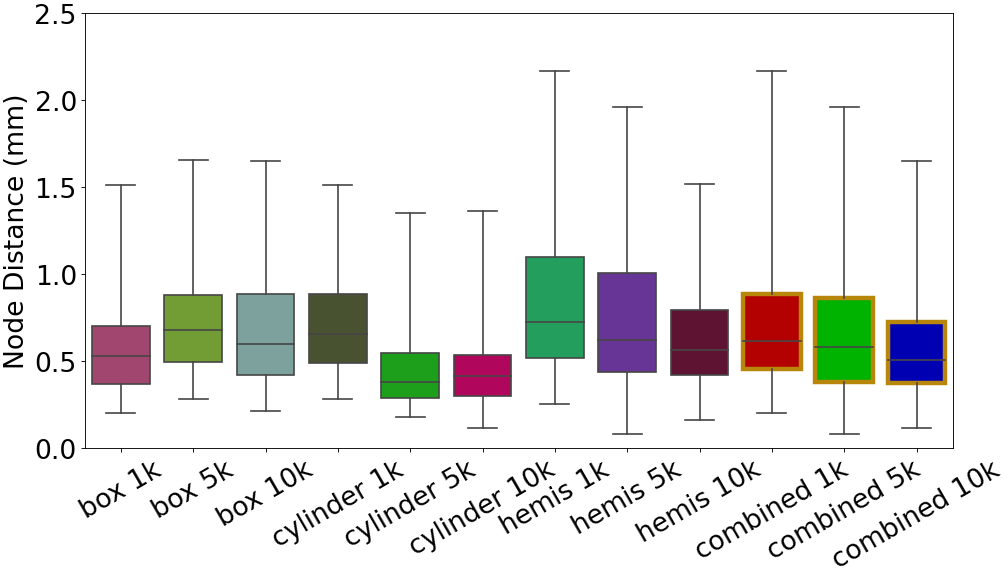} \\
    \includegraphics[width=1\linewidth]{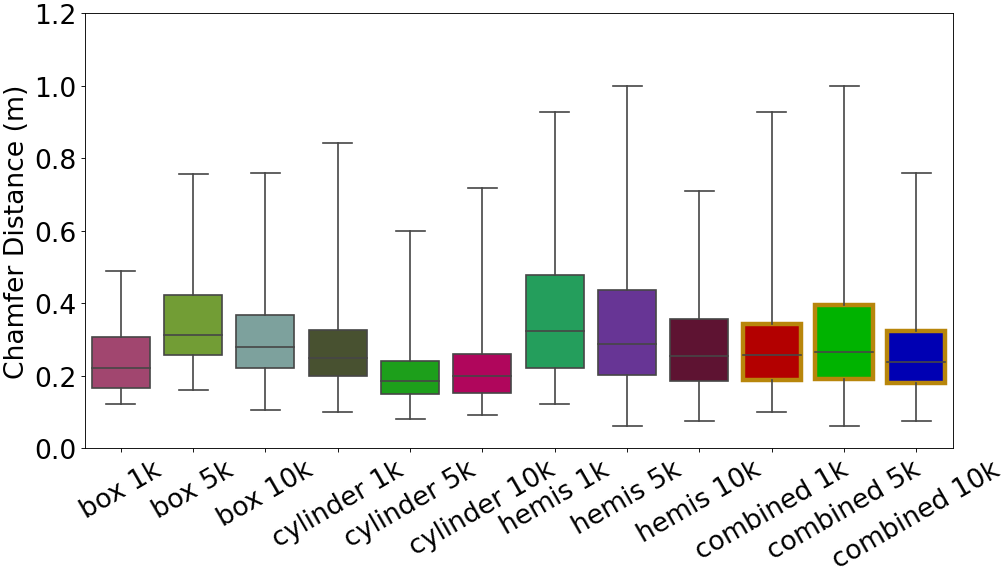} \\

    \caption{Experimental results for the single-arm manipulation case in simulation across the nine object categories. Each box-and-whiskers, corresponding to a specific object category, contains 100 data points obtained from evaluating our method on the 100 test goal shapes. The box represents the quartiles, the center line represents the median, and the whiskers represent the min and max final node/Chamfer distance. The last three box-and-whiskers with gold-color edges are aggregate results obtained from all object categories with the same stiffness range. \textit{(Top)} Node distance results. \textit{(Bottom)} Chamfer distance results.}

    \label{fig:single_sim_exp_results}

\end{figure}

Fig.~\ref{fig:single_sim_exp_results} presents the node and Chamfer distance results for each of the 9 object categories. Each box-and-whiskers, corresponding to a specific object category, contains 100 data points obtained from evaluating our method on the 100 test goal shapes. 
The box represents the quartiles, the center line represents the median, and the whiskers represent the min and max final node/Chamfer distance.

\revised{With respect to the number of steps to shape servoing convergence, all nine object categories exhibit fairly consistent results. Specifically, for the box primitive, \DeformerNet{} on average requires 1.5, 1.4, and 1.8 steps in the \unit[1]{kPa}, \unit[5]{kPa}, and \unit[10]{kPa} categories, respectively. Similarly, for the cylinder primitive, our method shows an average step count of 1.4, 1.6, and 1.5 steps. When dealing with the hemi-ellipsoid primitive, our method requires an average of 1.5, 1.9, and 1.8 steps. These results highlight the efficiency of \DeformerNet{}, as it consistently accomplishes the shape servoing task with a reasonably small number of steps.}

\begin{figure*}[t!]
    \centering
    \includegraphics[width=1.0\textwidth]{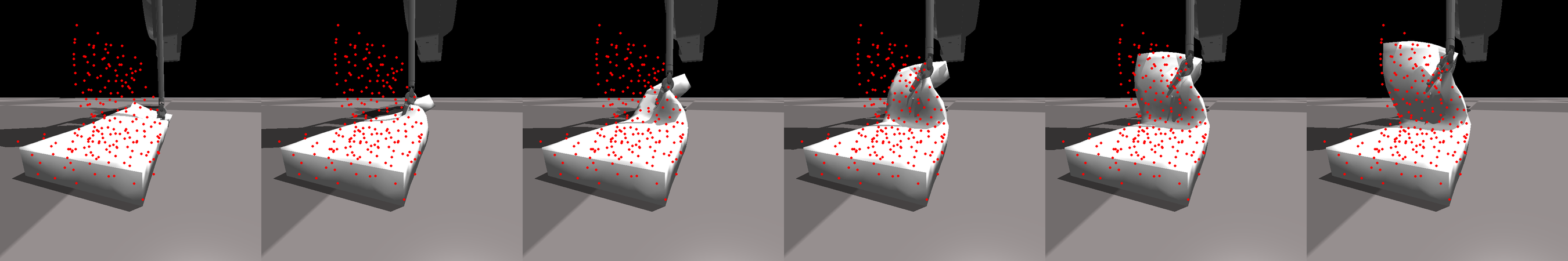} 
    \caption{Sample manipulation sequence of single-arm \DeformerNet{} with the simulated dVRK in Isaac Gym ($0.505$\,mm node distance and $0.252$\,m Chamfer distance).}
    \label{fig:sequence_single_sim}
\end{figure*}

Node/Chamfer distance by themselves do not provide an intuitive and qualitative understanding of how well our method performs on the test goal shapes.
Therefore in Fig.~\ref{fig:sequence_single_sim}, we provide a sample manipulation sequence of the robot performing shape servoing to a goal shape. More example manipulation sequences are provided in the supplementary video attachment.
Additionally in Fig.~\ref{fig:single_quartiles_sim}, we select interesting test goal point clouds and visualize the final object shapes after running our framework with them.
Specifically, we look at all the data points from evaluating the box primitive, and visualize the final shapes at the minimum (best result), $25^\textrm{th}$ percentile, median, $75^\textrm{th}$ percentile, and maximum (worst result) data points.
The visual results show that even at the maximum (worst case), the final shape still looks decently similar to the goal.
From the $75^\textrm{th}$ percentile to the minimum, our method all qualitatively succeeds in matching the goal shape.

\begin{figure*}[t!]
    \centering
    \includegraphics[width=1.0\textwidth]{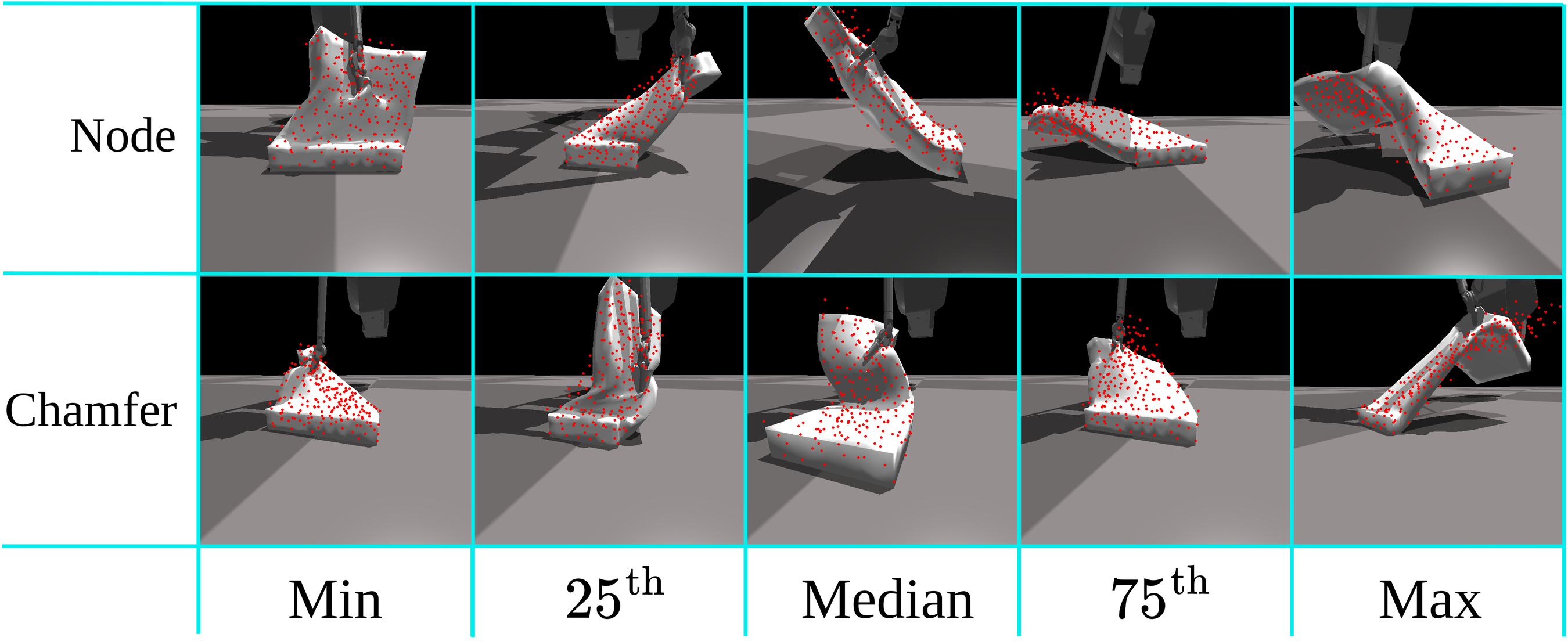}
    \vspace{0pt}
    \caption{Final object shapes of the box primitive (and the corresponding goal point clouds visualized in red) at the minimum, $25^\textrm{th}$ percentile, median, $75^\textrm{th}$ percentile, and maximum data points, from left to right respectively - \textbf{Single-arm case, in simulation}.\\
    \textit{(Top row)} With respect to the node distance evaluation metric. Node distances from left to right: $0.280, 0.404, 0.592, 0.871,$ and $1.655$\,mm. Chamfer distances from left to right: $0.156, 0.129, 0.234, 0.313,$ and $0.323$\,m.\\
    \textit{(Bottom row)} With respect to the Chamfer distance evaluation metric. Node distances from left to right: $0.221, 0.356, 0.505, 0.764,$ and $1.26$\,mm. Chamfer distances from left to right: $0.139, 0.198, 0.252, 0.371,$ and $0.757$\,m.} 
    \label{fig:single_quartiles_sim}
\end{figure*}

\subsubsection{Manipulation point selection}\label{sec:mani_point_selection}

\revised{We compare the performance of $\DeformerNet{}$ when using our primary manipulation point selection method, \textit{dense predictor}, against two alternatives (\textit{classifier} and \textit{keypoint-based heuristic}) as well as an oracle. \textit{Oracle} in this context refers to the ground-truth manipulation points used when generating the test goal shapes and can be viewed as the best possible manipulation points the robot can choose.}

\revised{The first competitive alternative to \textit{dense predictor} is the \textit{classifier}, also a very popular technique in the robot grasping community~\cite{mousavian2019graspnet,lu-ram2020-grasp-inference,lu-iros2020-active-grasp, vandermerwe-icra2020-reconstruction-grasping}. We have extended this approach to deformable object manipulation by designing a neural network very similar to $\DeformerNet{}$; the only difference is that the output is modified to produce a scalar value. The model takes as inputs the current and goal point clouds, as well as a candidate manipulation point. It outputs the likelihood of this candidate being a good manipulation point (normalized to lie between 0 and 1 using the sigmoid function). At runtime, we sample a set of $N$ candidates and evaluate their likelihoods. The manipulation point can then be straightforwardly defined as the candidate with the highest likelihood.} 

\revised{The second alternative is the \textit{keypoint-based heuristic} method from our previous work~\cite{thach2022learning}.}

\begin{figure}[ht]
    \centering
    \includegraphics[width=1\linewidth]{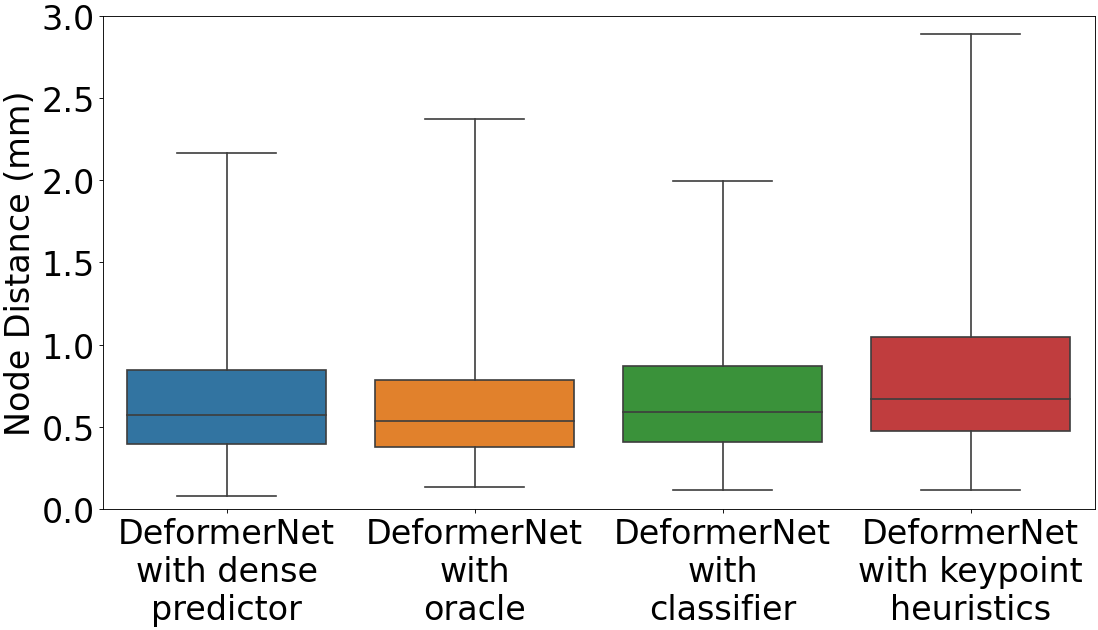} \\
    \includegraphics[width=1\linewidth]{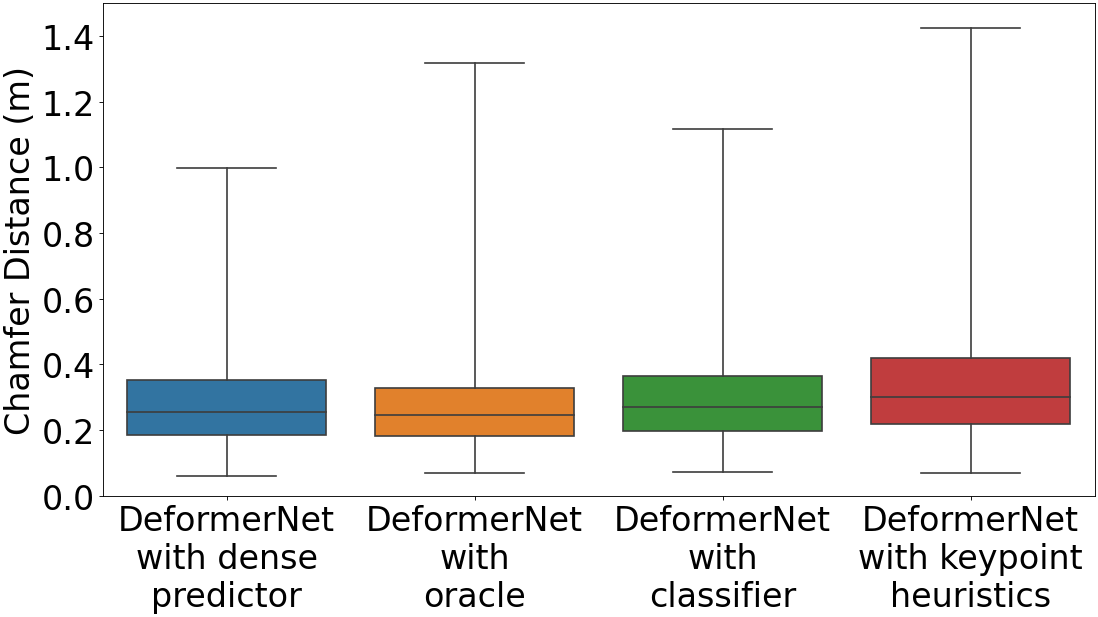} \\    
    \caption{Distribution of node distance and Chamfer distance when using different manipulation point selection techniques.}
    \label{fig:ablation_mp}
\end{figure}

\revised{To evaluate the manipulation point selection methods we run the same experiments as in Sec.~\ref{sec:generalization_performance} above, but combine the results from all nine object categories together for plotting. 
As can be seen in Fig.~\ref{fig:ablation_mp}, our \textit{dense predictor} performs on par with \textit{oracle} and \textit{classifier}, while outperforming \textit{keypoint-based heristic} by a substantial margin. It is worth noting that, 1) at runtime the robot does not have access to the \textit{oracle} manipulation points, and 2) the \textit{classifier} runs much slower than the \textit{dense predictor} as it requires many forward passes through the network to evaluate multiple manipulation point candidates, while \textit{dense predictor} requires only a single forward pass.}

\revised{Evaluating steps to convergence, when combined with \DeformerNet{}, the \textit{dense predictor} on average requires 1.6 steps to finish the shape servoing task, which shows comparable performance to the average of 1.5 steps required by the \textit{oracle}. 
The \textit{classifier} method requires an average step count of 1.6, demonstrating similar efficiency to the \textit{dense predictor}. The \textit{keypoint-based heuristic} yields the worst result of 2.4 steps.}

\subsubsection{Ablation study 1 - \DeformerNet{} with vs without manipulation point as input}
Beyond the conference paper version of this work~\cite{thach2022learning}, our upgraded \DeformerNet{} also takes the selected manipulation point as an input. We conduct the same experiments as Sec.~\ref{sec:generalization_performance}, but with models that are trained without the manipulation point information. Fig.~\ref{fig:ablation_architecture} (second box-and-whiskers) visualizes our results. We can observe a substantial decrease in performance when the manipulation point location is hidden from \DeformerNet{}.
\revised{With respect to the number of steps to convergence, this ablated version of \DeformerNet{} consumes on average 1.8 steps to finish the shape servoing task, demonstrating a reduction in performance as compared to the average 1.6 steps achieved by the full \DeformerNet{}.}

\subsubsection{Ablation study 2 - \DeformerNet{} with vs without orientation}
\begin{figure}[ht]
    \centering

    \includegraphics[width=1\linewidth]{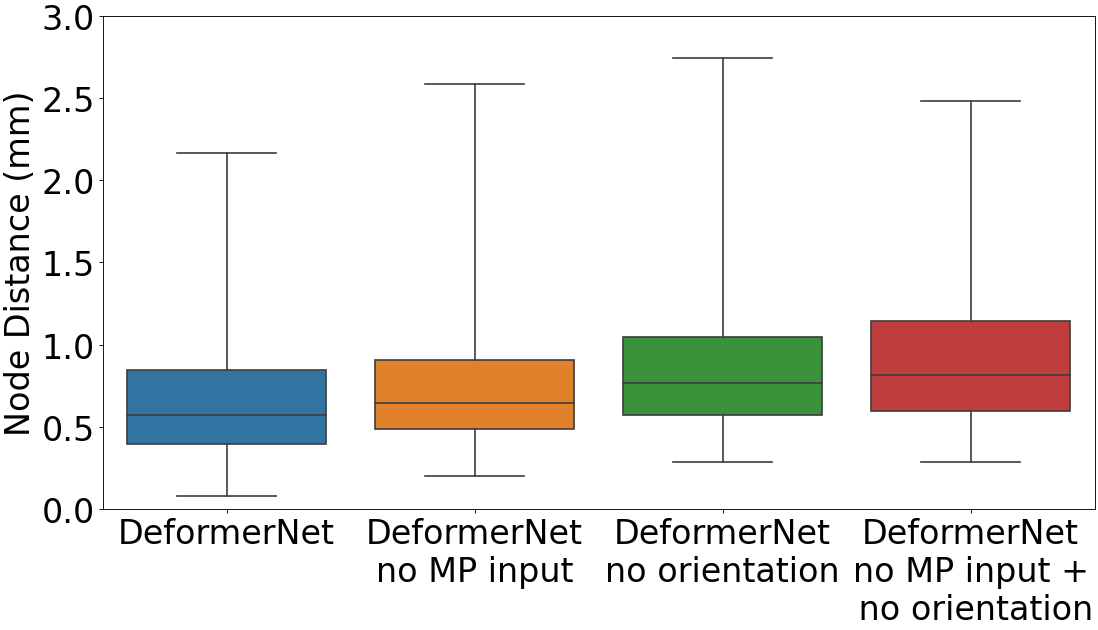} \\
    \includegraphics[width=1\linewidth]{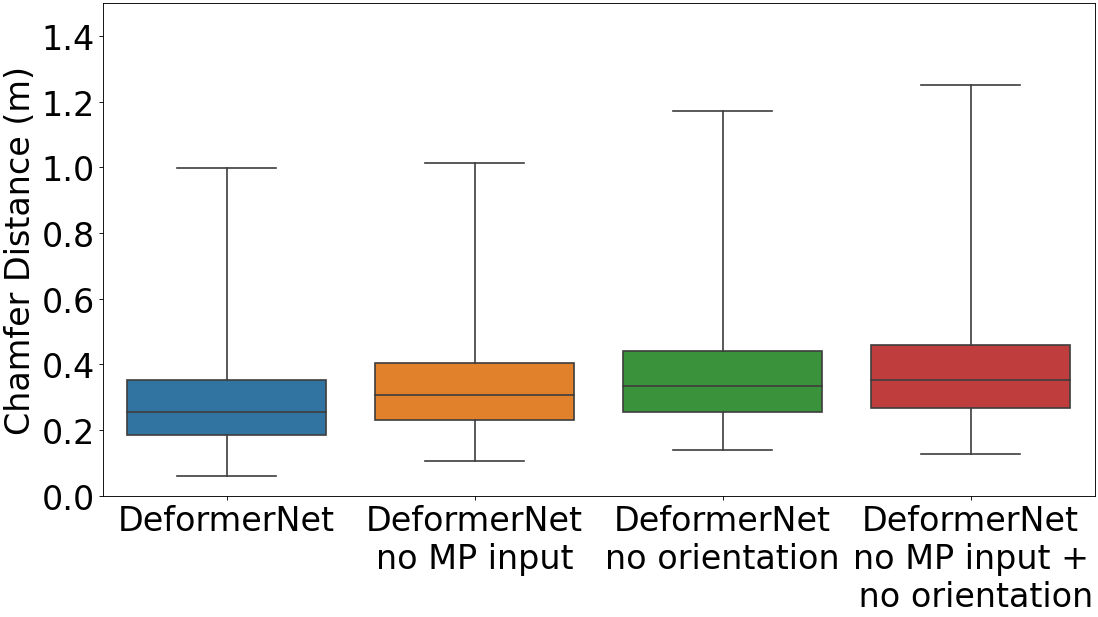} \\
    \caption{Ablation study - Distribution of node distance and Chamfer distance if we remove specific features from the architecture of \DeformerNet{}.}
    \label{fig:ablation_architecture}
\end{figure}

Unlike the conference paper version of our work~\cite{thach2022learning} which limits the action space to only gripper position, our upgraded \DeformerNet{} architecture enables the robot to apply a change in both position and orientation of its end-effector. In this section, we evaluate the effectiveness of this new contribution. We conduct the same experiments as~\ref{sec:generalization_performance}, but using our previously trained models~\cite{thach2022learning} which only output gripper position displacement. Fig.~\ref{fig:ablation_architecture} (third box-and-whiskers) visualizes our results. \DeformerNet{} with the expanded action space leads to better performance, most-likely because it is more expressive and enables the deformable objects to reach more complex shapes. Furthermore, when we remove from the \DeformerNet{} architecture both the orientation displacement in the action space and the manipulation point input, we observe a worse performance than when removing just either of those features (Fig.~\ref{fig:ablation_architecture} last box-and-whiskers).
\revised{With respect to the number of steps to convergence, these two ablated versions of \DeformerNet{} require on average 2.0 and 2.5 steps, respectively, demonstrating a noticeable reduction in performance as compared to the average 1.6 steps achieved by the full \DeformerNet{}.}

\subsubsection{Comparison with other planning methods}
We also compare the performance of our method against Rapidly-exploring Random Tree (RRT) \cite{lavalle2001randomized} and model-free Reinforcement Learning (RL) for the 3D shape servo problem.
Here we restrict the task to be trained and tested on a single box object and use only one manipulation point throughout training and testing.

For the RRT implementation, we define the configuration space as the joint angles of the dVRK manipulator.
We define a goal region as any object point cloud that has Chamfer distance less than some tolerance from the goal point cloud.
We use the finite element analysis model~\cite{macklin2019} in the Isaac Gym~\cite{Liang2018GPU} simulator to derive the forward model for RRT.

We use proximal policy optimization (PPO)~\cite{schulman2017proximal} (as in~\cite{pore2021safe}) with hindsight experience replay (HER)~\cite{andrychowicz2017hindsight} for model-free RL.
We use our \DeformerNet{} architecture for the actor and critic network except for the critic output being set to a single scalar to encode the value function.
In each episode, we condition the policy on a newly sampled goal shape.
We train the RL agent with 200,000 samples---10 times the amount of data provided to \DeformerNet{}.

\begin{figure}[th]
    \centering
    \includegraphics[width=1\linewidth]{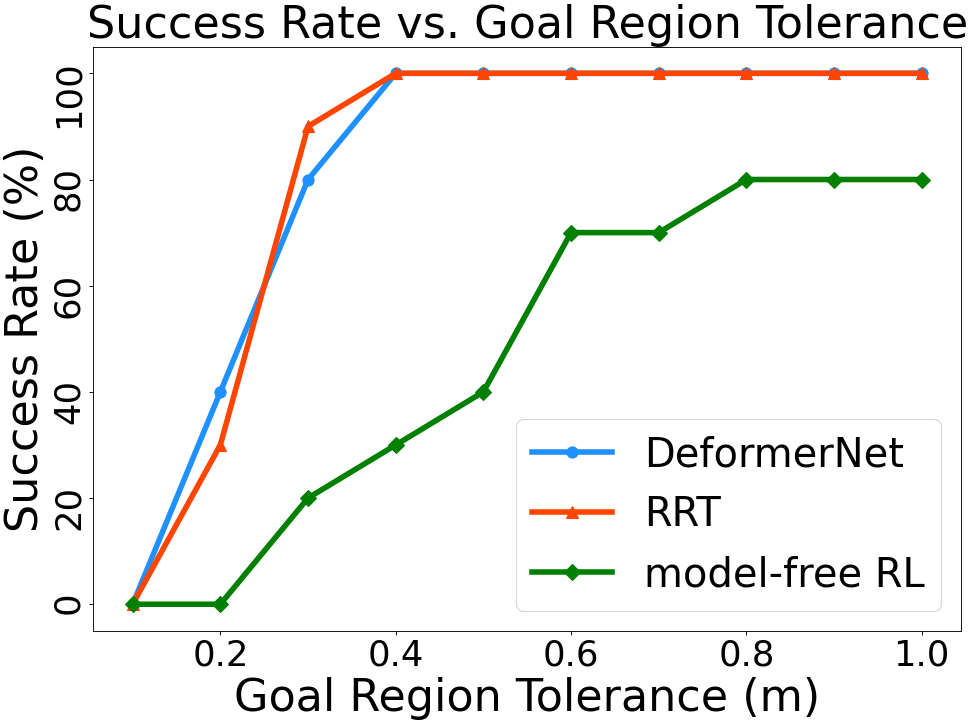}
    \caption{Success rate comparison of \DeformerNet{} to RRT and model-free RL for varying levels of goal tolerance (defined as Chamfer distance in meters).}
    \label{fig:success_rate_baseline}
\end{figure}

We evaluate DeformerNet, RRT, and model-free RL with 10 random goal shapes.
Fig.~\ref{fig:success_rate_baseline} shows the success rate of the three methods at different levels of goal tolerance.
We clearly see that even with 10 times the training data compared to our method, the model-free RL agent achieves a significantly lower success rate compared to the other two methods. 
We also note that, even though RRT performs comparably to our method, it has some critical shortcomings.
Unlike our method, RRT cannot incorporate feedback during execution. As a result, RRT will not be able to recover if the object shape deviates from the plan. While one might think to perform replanning, we note that RRT requires several orders of magnitude more computation time than our shape servoing approach. This is due to the fact that planning with RRT requires a forward model of the deformable object, which typically involves expensive Finite Element Method (FEM) computation.  
For instance, at a tolerance of 0.4 (where both our method and RRT achieve 100\% success), over the 10 test goal shapes, the lowest computation time required by RRT was 1.1 minutes, the highest was 110.32 minutes, mean was 25.25 minutes, and standard deviation was 32.27 minutes. Our \DeformerNet{}, however, only requires a pass through the neural network which takes minimal time.
As a result, for this task, we note a significant success rate improvement for our method over model-free RL and a significant computation time improvement over RRT in all cases.

\subsubsection{Bimanual manipulation}\label{sec:bimanual_manipulation} 
\begin{figure}[ht]
    \centering
    \includegraphics[width=1\linewidth]{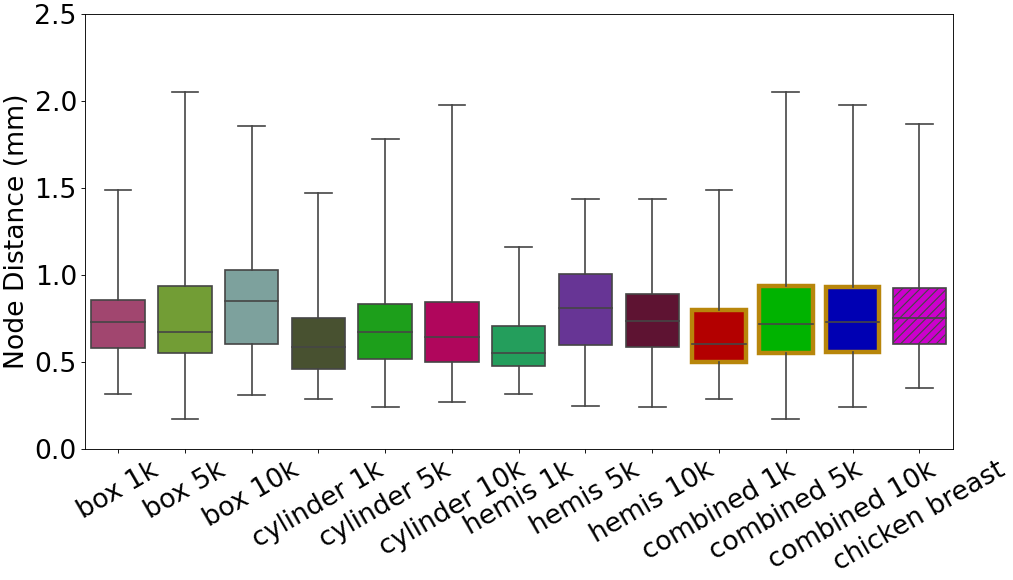} \\
    \includegraphics[width=1\linewidth]{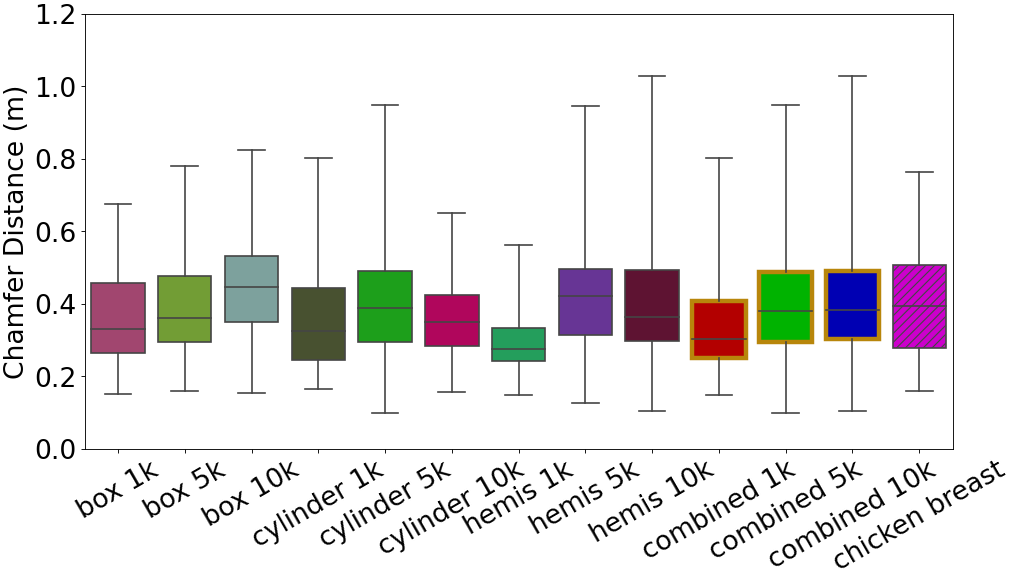} \\ 
    \caption{Experimental results for the bimanual manipulation case in simulation across the nine object categories. Each box-and-whiskers, corresponding to a specific object category, contains 100 data points obtained from evaluating our method on the 100 test goal shapes. The box represents the quartiles, the center line represents the median, and the whiskers represent the min and max final node/Chamfer distance. The three box-and-whiskers with gold-color edges are aggregate results obtained from all object categories with the same stiffness range. 
    The last box-and-whisker, distinguished by the forward slash hatch pattern, represents the experimental results for the chicken breast.
    \textit{(Top)} Node distance results. \textit{(Bottom)} Chamfer distance results.}
    \vspace{-10pt}
    \label{fig:bimanual_sim_exp_results}
\end{figure}

$\DeformerNet$ opens the door to many applications where more than one robot arm is required to accomplish a task.
Here we evaluate the full bimanual version of \DeformerNet{} in simulation, similarly to the above single-arm section.
The boxplot results for each of the 9 object categories are presented in Fig.~\ref{fig:bimanual_sim_exp_results}. Each box-and-whiskers, corresponding to a specific object category, contains 100 data points obtained from evaluating our method on the 100 test goal shapes. 
The box represents the quartiles, the center line represents the median, and the whiskers represent the min and max final node/Chamfer distance.

\begin{figure*}[t!]
    \centering
    \includegraphics[width=1.0\textwidth]{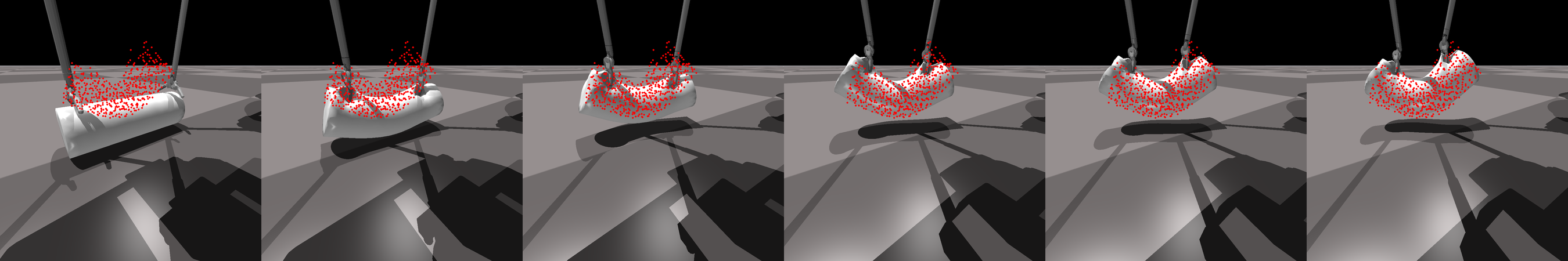}  
    
    \caption{Sample manipulation sequence of the bimanual version of \DeformerNet{} with simulated dVRK in Isaac Gym ($0.50$\,mm node distance and $0.27$\,m Chamfer distance).} 
    \label{fig:sequence_bimanual_sim}
\end{figure*}

\begin{figure*}[t!]
    \centering
    \includegraphics[width=1.0\textwidth]{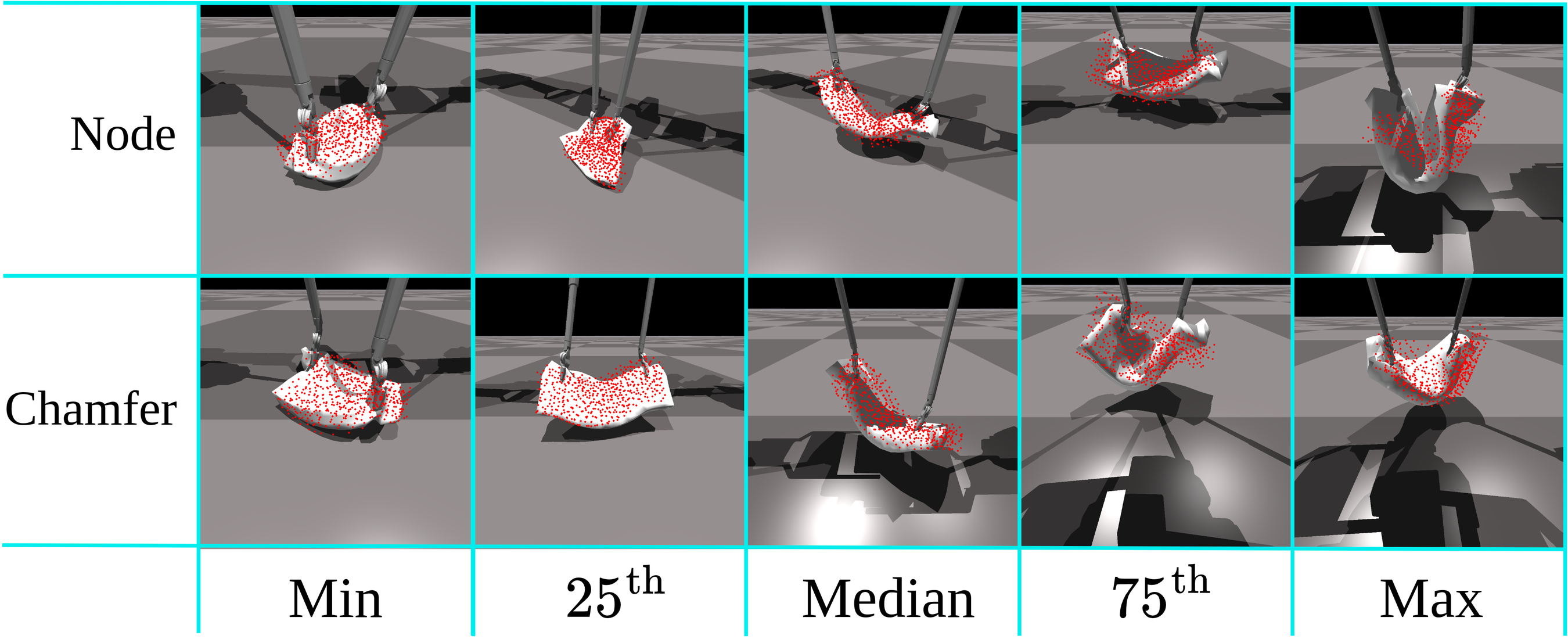}  
    \caption{Final object shapes of the box primitive (and the corresponding goal point clouds visualized in red) at the minimum, $25^\textrm{th}$ percentile, median, $75^\textrm{th}$ percentile, and maximum data points, from left to right respectively - \textbf{Bimanual manipulation case, in simulation}. \\
    \textit{(Top row)} With respect to the node distance evaluation metric. Node distances from left to right: $0.229, 0.564, 0.730, 0.958,$ and $2.050$\,mm. Chamfer distances from left to right: $0.168, 0.222, 0.223, 0.559,$ and $0.674$\,m. \\
    \textit{(Bottom row)} With respect to the Chamfer distance evaluation metric. Node distances from left to right: $0.331, 0.630, 0.977, 0.921,$ and $2.022$\,mm. Chamfer distances from left to right: $0.142, 0.291, 0.383, 0.494,$ and $0.782$\,m.} 
    \label{fig:bimanual_quartiles_sim}
\end{figure*}

Similarly to the single-arm case, in Fig.~\ref{fig:sequence_bimanual_sim}, we present a sample snapshot of the dual-arm robot successfully performing shape servoing to a goal shape. More example manipulation sequences are provided in the supplementary video attachment.
Additionally, to provide an intuitive and qualitative understanding of how well our method performs on the test goal shapes, we look at all the data points from evaluating the box primitive and visualize the data points at the minimum (best result), $25^\textrm{th}$ percentile, median, $75^\textrm{th}$th percentile, and maximum (worst result). Please refer to Fig.~\ref{fig:bimanual_quartiles_sim} for this qualitative visualization.
The visual results show that even at the maximum (worst result), the final shape still looks decently similar to the goal.
From the $75^\textrm{th}$ percentile to the minimum, our method all qualitatively succeeds in matching the goal shape. 

\revised{With respect to the number of steps metric, all nine object categories exhibit fairly consistent results. Specifically, for the box primitive, \DeformerNet{} on average requires 2.2, 2.3, and 2.1 steps in the \unit[1]{kPa}, \unit[5]{kPa}, and \unit[10]{kPa} categories, respectively. Similarly, for the cylinder primitive, our method requires an average step count of 2.7, 2.4, and 2.6 steps. For the hemi-ellipsoid primitive, our method requires an average of 2.2, 2.3, and 2.6 steps.}

\begin{figure*}[t!]
    \centering
    \includegraphics[width=1.0\textwidth]{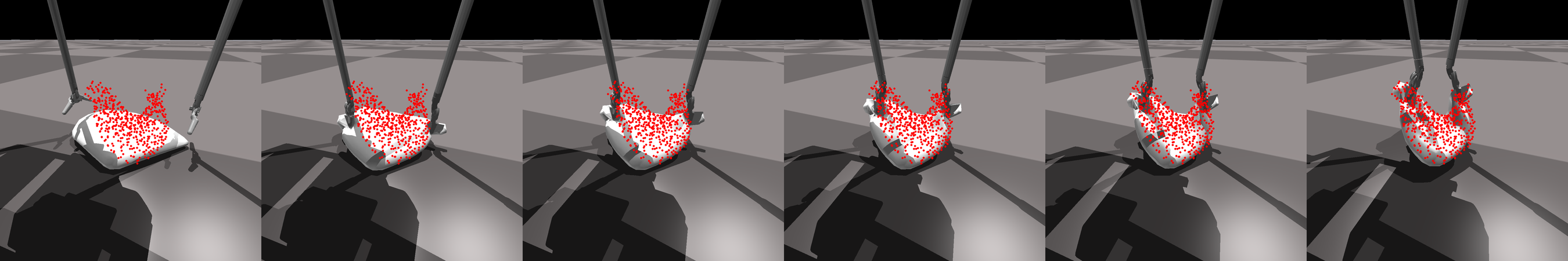}  
    \caption{Sample manipulation sequence with the chicken breast, an object with complex geometry that was unseen during training, in Isaac Gym ($0.358$\,mm node distance and $0.201$\,m Chamfer distance).} 
    \label{fig:sequence_bimanual_sim_chicken}
\end{figure*}

\begin{figure*}[t!]
    \centering \includegraphics[width=1.0\textwidth]{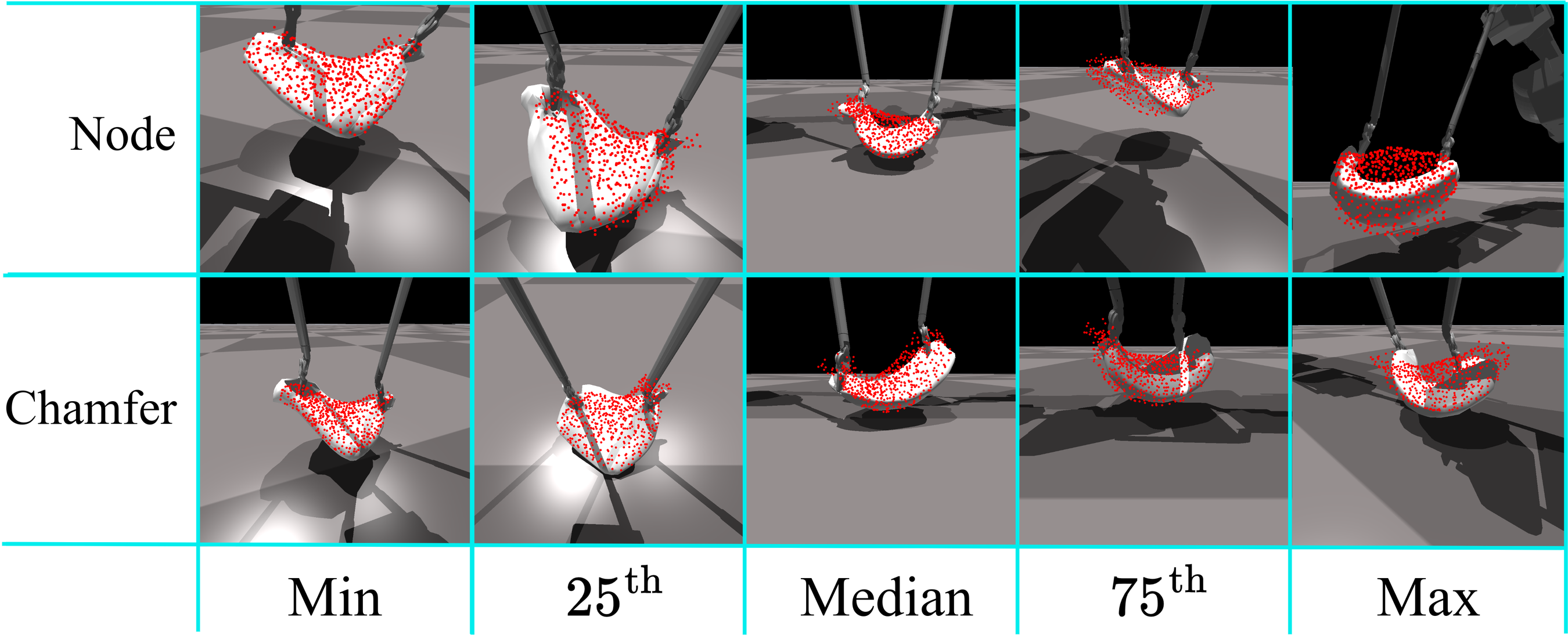}  
    \caption{\revised{Final object shapes of the chicken breast (and the corresponding goal point clouds visualized in red) at the minimum, $25^\textrm{th}$ percentile, median, $75^\textrm{th}$ percentile, and maximum error data points, from left to right respectively - \textbf{Bimanual manipulation case, in simulation}. \\
    \textit{(Top row)} With respect to the node distance evaluation metric. Node distances from left to right: $0.350, 0.591, 0.801, 1.02,$ and $1.866$\,mm. Chamfer distances from left to right: $0.195,0.223,0.303,0.435,$ and $0.550$\,m. \\
    \textit{(Bottom row)} With respect to the Chamfer distance evaluation metric. Node distances from left to right: $0.516,0.803,0.680,1.141,$ and $1.026$\,mm. Chamfer distances from left to right: $0.193,0.301,0.381,0.502,$ and $0.754$\,m.}} 
    \label{fig:bimanual_quartiles_sim_chicken}
\end{figure*}

\subsubsection{\revised{Performance on a complex, unseen object}}\label{sec:unseen_objects} 
\revised{To further demonstrate the generalizability of \DeformerNet{}, we challenge our shape servoing pipeline with bimanually manipulating a chicken breast (visualized in Fig.~\ref{fig:display_objects}), an object with complex geometry that was not only \textit{unseen} during training but also outside the training distribution. 
To facilitate this evaluation, we utilize \DeformerNet{} with the exact same architecture as before (Fig.~\ref{fig:bimanual_DeformerNet}). However, instead of having a separate model for each primitive as in Sec.~\ref{sec:bimanual_manipulation}, we train this new model on a meta dataset merging all data collected in Sec.~\ref{sec:training_data_generation} together (including all box, cylinder, and hemi-ellipsoid primitives).}

\revised{We evaluate the performance of our method on 100 instances of the chicken breast in simulation, each characterized by a unique stiffness uniformly sampled from \unit[1]{kPa} to \unit[10]{kPa}. We evaluate on 100 test goal shapes corresponding to each of these instances. As illustrated in Fig.~\ref{fig:bimanual_sim_exp_results}, the results obtained on this complex object (rightmost box-and-whisker) remain comparable to those achieved on the box, hemi-ellipsoid, and cylinder primitives. The qualitative results in Fig.~\ref{fig:sequence_bimanual_sim_chicken} and Fig.~\ref{fig:bimanual_quartiles_sim_chicken} also show that our method succeeds in matching the challenging goal shapes of the chicken breast.
With respect to the step count metric, our method requires on average 2.6 steps to complete the shape servoing task, achieving similar performance to those observed in the box, hemi-ellipsoid, and cylinder primitives.}

\subsection{Goal-Oriented Shape Servoing on the Physical Robot}
We next evaluate our method's ability to perform shape servoing on a physical robot, \revised{while having been trained entirely in simulation on the dVRK manipulator arms as described above}.
The experimental setup is shown in Fig.~\ref{fig:exp_setup}. The robot is tasked with manipulating a box-shaped deformable object and a cylindrical tube to several goal shapes, both in the single-arm and bimanual manipulation cases. For the single-arm setup, the manipulated objects are affixed on one side to a table. \revised{We remind the reader that \DeformerNet{} outputs homogeneous end-effector transforms which are translated into joint velocities by our resolved-rate controller. These are then executed via the existing joint-level controller of the Baxter robot without any need for fine-tuning.}

We segment the object's point cloud out from the rest of the scene by first excluding points that are too far away from the object, and then filtering out the object using pixel intensity. \revised{Specifically, we define a bounding box surrounding the object, excluding points outside the bounding box. As segmentation is not the focus of this work, we leverage distinctive colors for the objects, ensuring a clear contrast in pixel intensities between the object and the background, surgical tool, and table. As a result, the surgical tools and the background point clouds do not interfere with the current point cloud.}

\begin{figure}[ht]
    \centering
    \includegraphics[width=1\linewidth]{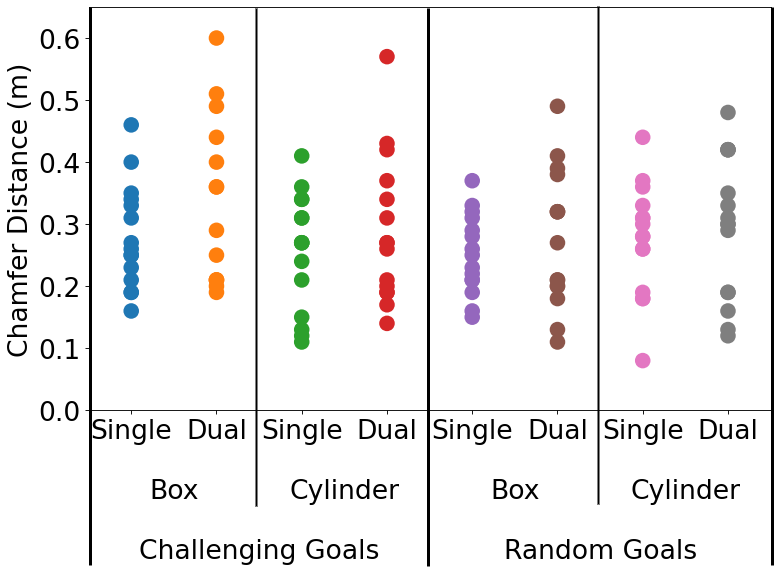}%
    \vspace{5pt}
    \caption{Physical robot experimental results. Distribution of Chamfer distance across various test scenarios, including two \DeformerNet{} versions (single-arm and dual-arm), two target objects (box-shaped deformable object and cylindrical tube), and two goal shape categories (challenging and random).}
    \label{fig:real_exp_results}
\end{figure}

\begin{figure*}[t!]
    \centering
    \includegraphics[width=1.0\textwidth]{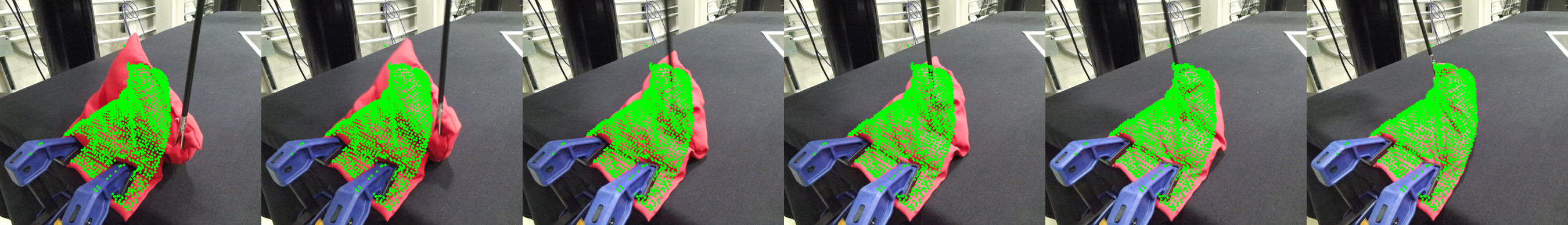}
    \includegraphics[width=1.0\textwidth]{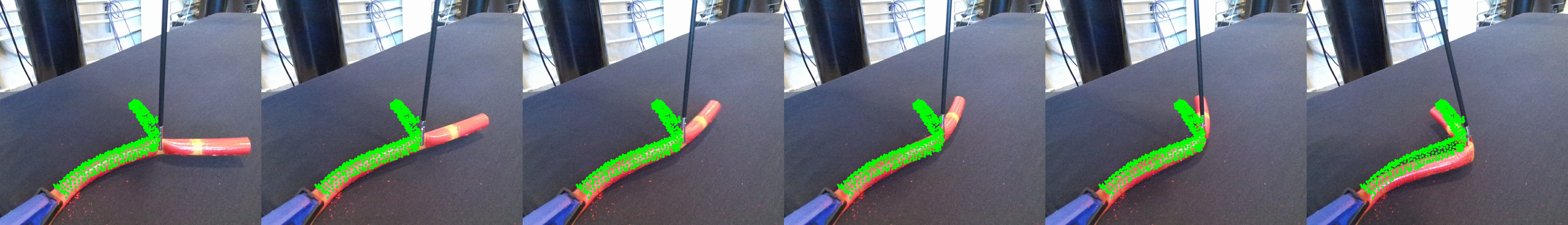}    
    \caption{Sample manipulation sequences of single-arm \DeformerNet{} with physical robot in different setups. First row: on a box-shaped pillow ($0.27$\,m final Chamfer dist). Second row: on a cylindrical tube ($0.34$\,m final Chamfer dist).}
    \label{fig:sequence_single_real}
\end{figure*}

\begin{figure*}[t!]
    \centering
    \includegraphics[width=1.0\textwidth]{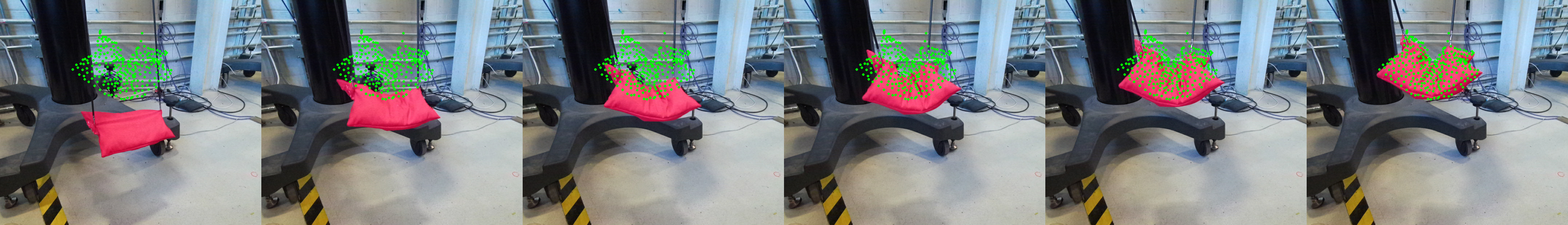}
    \includegraphics[width=1.0\textwidth]{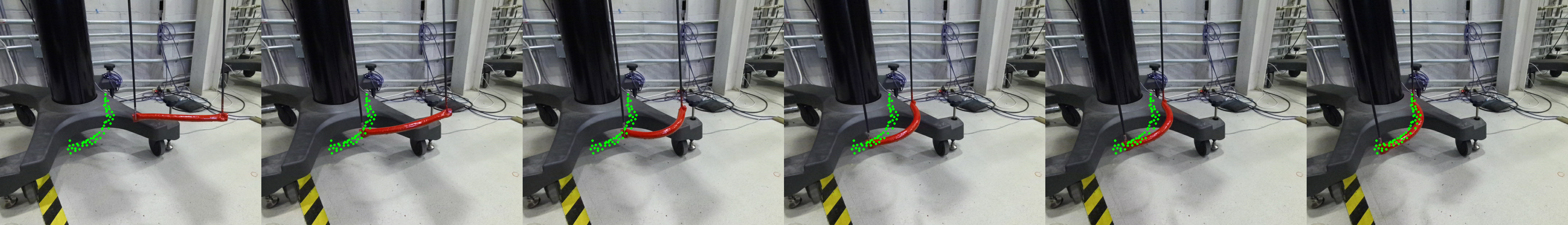}    
    \includegraphics[width=1.0\textwidth]{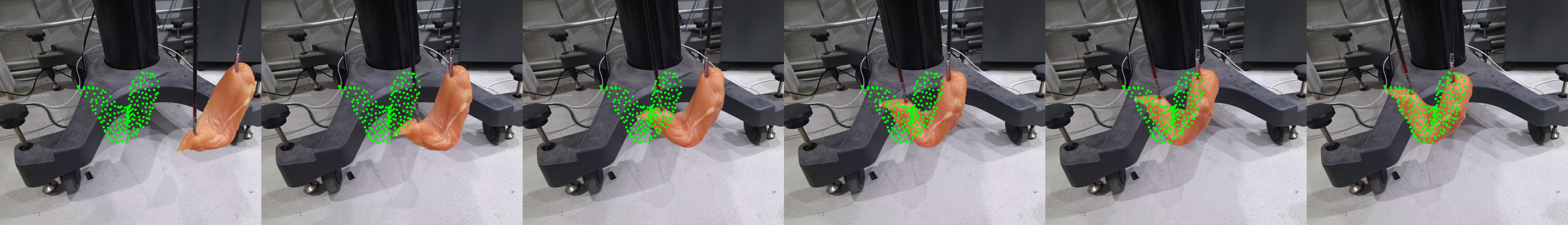} 
    \caption{Sample manipulation sequences of the bimanual version of \DeformerNet{} with physical robot in different setups. First row: on a box-shaped object ($0.25$\,m final Chamfer dist). Second row: on a cylindrical tube ($0.34$\,m final Chamfer dist). \revised{Third row: on ex vivo chicken muscle tissue ($0.26$\,m final Chamfer dist).}} 
    \label{fig:sequence_bimanual_real}
\end{figure*}

\begin{figure*}[t!]
    \centering
    \includegraphics[width=1.0\textwidth]{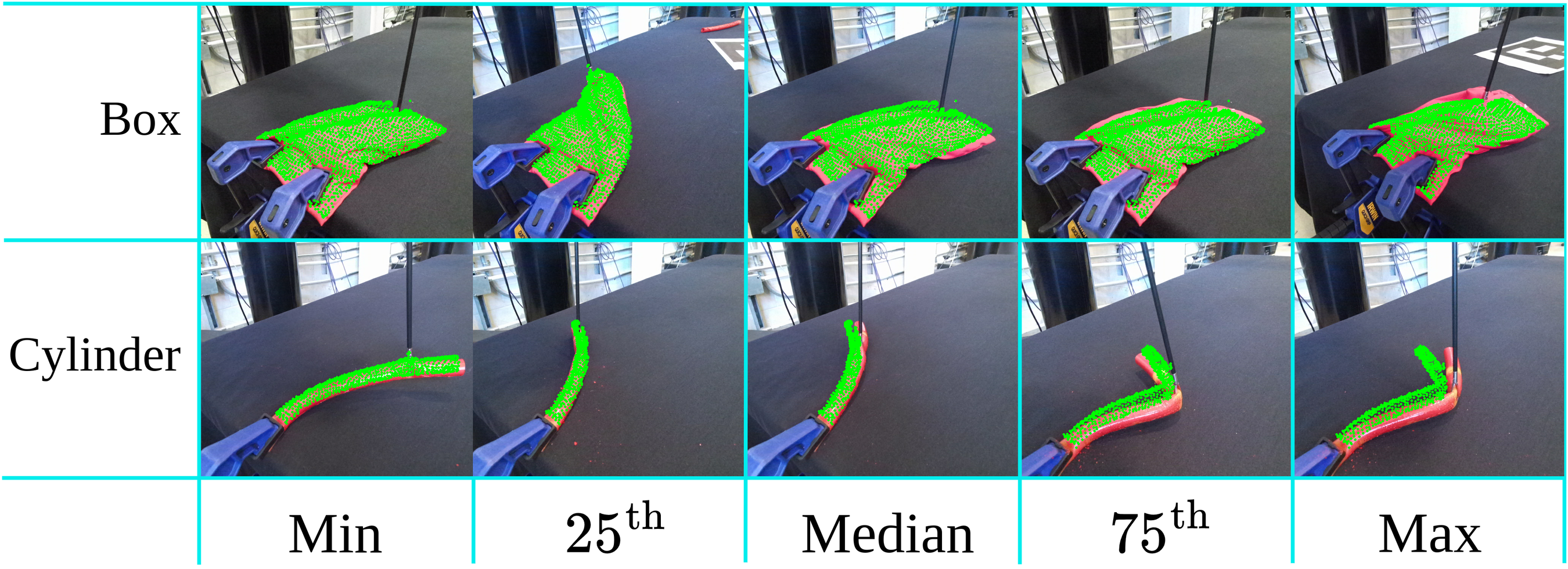} 
    \caption{Final object shapes (and the corresponding goal point clouds visualized in green) at the minimum, $25^\textrm{th}$ percentile, median, $75^\textrm{th}$ percentile, and maximum data points, from left to right respectively - \textbf{Single-arm case, on the physical robot}.\\ 
    \textit{(Top row)} Box-like deformable object; Chamfer distances from left to right: $0.19, 0.25, 0.26, 0.35,$ and $0.46$\,m. 
    \textit{(Bottom row)} Cylindrical tube object; Chamfer distances from left to right: $0.12, 0.21, 0.27, 0.34,$ and $0.41$\,m.} 
    \label{fig:single_quartiles_real}
\end{figure*}

\begin{figure*}[t!]
    \centering
    \includegraphics[width=1.0\textwidth]{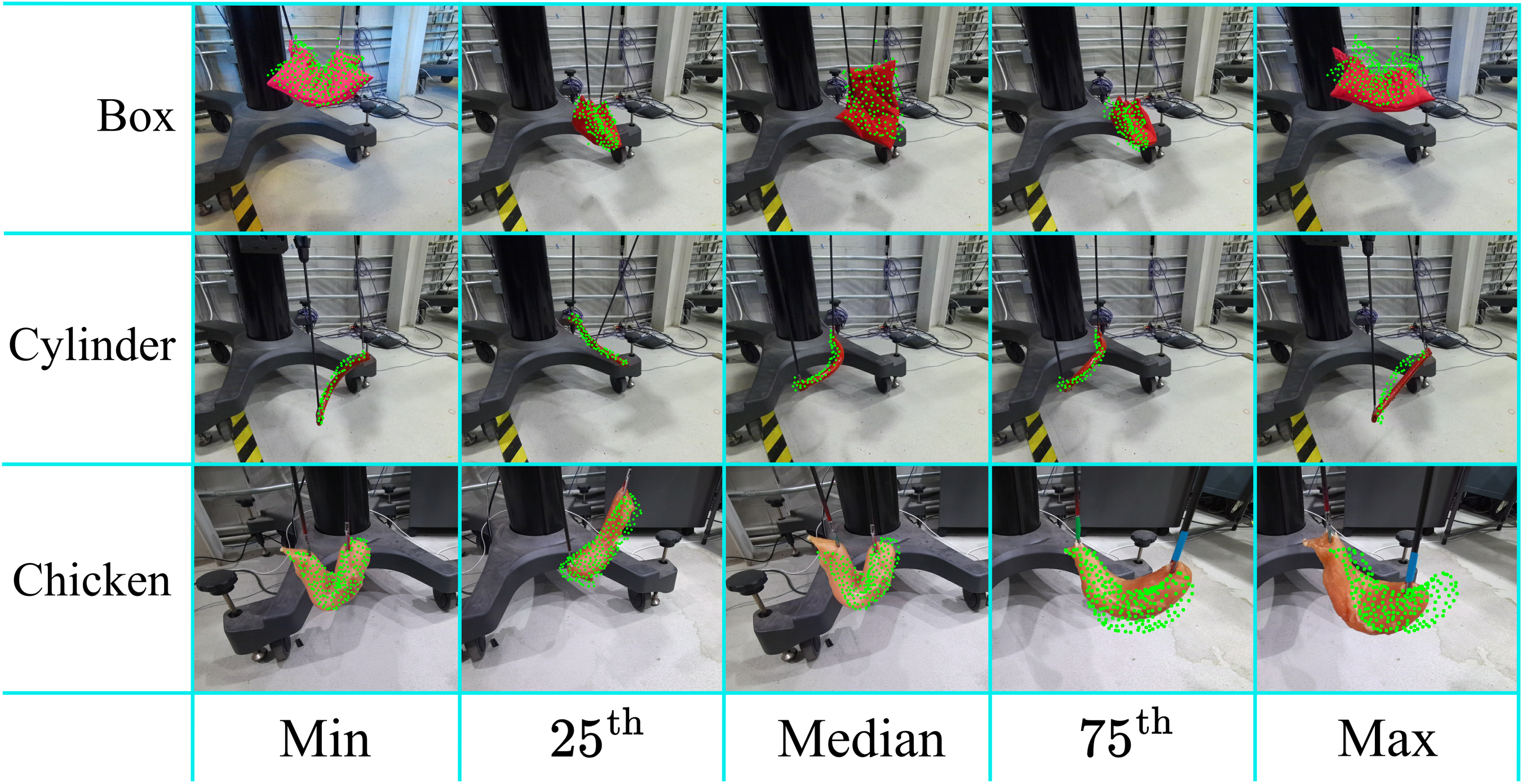}    
    \vspace{1pt}
    \caption{Final object shapes (and the corresponding goal point clouds visualized in green) at the minimum, $25^\textrm{th}$ percentile, median, $75^\textrm{th}$ percentile, and maximum data points, from left to right respectively - \textbf{Bimanual manipulation case, on the physical robot}.\\
    \textit{(Top row)} Box-like deformable object, with respect to the Chamfer distance evaluation metric. Chamfer distances from left to right: $0.19, 0.21, 0.36, 0.44,$ and $0.60$\,m.
    \textit{(Middle row)} Cylindrical tube object, with respect to the Chamfer distance evaluation metric. Chamfer distances from left to right: $0.17, 0.20, 0.31, 0.43,$ and $0.57$\,m.
    \revised{\textit{(Bottom row)} Ex vivo chicken muscle tissue, with respect to the Chamfer distance evaluation metric. Chamfer distances from left to right: $0.089, 0.206, 0.308, 0.411,$ and $0.683$\,m.}} 
    \label{fig:bimanual_quartiles_real}
\end{figure*}

For each of the two target objects (the box-shaped deformable object and the cylindrical tube), we first generate three challenging goal shapes by manually moving the robot arm by hand. These goals are challenging because we control the arm such that the final end-effector pose has a large position and orientation displacement from the home position, hence ensuring the final object shapes are interesting and not easy to reach.
We then generate three random goal shapes by applying random actions to the robot. The robot performs 5 shape servoing trials on each goal shape, starting with 5 distinct initial shapes that are substantially different from the goal shape.
\revised{In all cases the actions that produced the goal shapes are discarded and only the recorded goal point clouds used for evaluation.}
Fig.~\ref{fig:real_exp_results} visualizes the results of eight test scenarios; each scatter plot consists of 15 data points from 3 goal shapes. Figure~\ref{fig:sequence_single_real} and~\ref{fig:sequence_bimanual_real} show a few sample manipulation sequences of the single-robot and bimanual cases, respectively. More manipulation sequences are provided in the supplementary video attachment. Overall, we observed that our shape servoing framework qualitatively succeeds in most test cases.
To better understand the local-minimum instances as well as to demonstrate that $\DeformerNet$ still yields decent results even for these cases,
in Fig.~\ref{fig:single_quartiles_real} and Fig.~\ref{fig:bimanual_quartiles_real} we visualize the final shapes at the minimum (best result), $25^\textrm{th}$ percentile, median, $75^\textrm{th}$ percentile, and maximum (worst result) data points, with respect to the final Chamfer distance.

\begin{figure}[ht]
    \centering
    \includegraphics[width=1\linewidth]{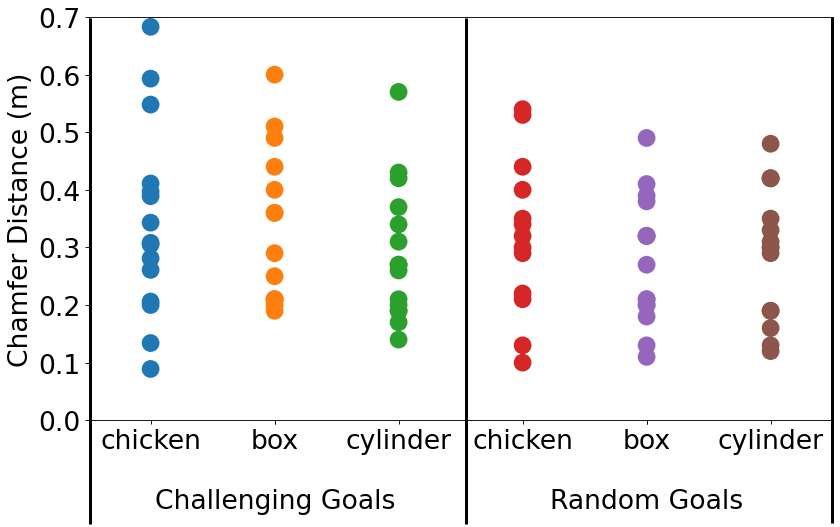}%
    \vspace{5pt}
    \caption{\revised{Physical robot experimental results on the bimanual manipulation of ex vivo chicken muscle tissue. We present the results of the ex vivo experiment alongside those of the box-shaped object and the cylindrical tube, highlighting that our method's performance remains comparable.}}
    \label{fig:real_exp_results_chicken}
    \vspace{-8mm}
\end{figure}

\revised{We also task the method with bimanually manipulating \textit{ex vivo} chicken muscle tissue on the physical robot. 
In Fig.~\ref{fig:real_exp_results_chicken}, we present the results of the \textit{ex vivo} chicken tissue experiment alongside those of the box-shaped object and the cylindrical tube, highlighting that our method's performance remains comparable on real tissue in this case. We present qualitative results of the chicken breast experiment in Fig.~\ref{fig:sequence_bimanual_real} and Fig.~\ref{fig:bimanual_quartiles_real}.} 

\revised{In terms of the step count metric, in the single-arm case, \DeformerNet{} on average requires 2.2 and 2.3 steps for the box-shaped object and the cylindrical tube, respectively. In the bimanual scenario, our method shows an average step count of 2.7 and 3.0 steps. During bimanual manipulation of the chicken tissue, our shape servoing pipeline requires an average of 2.9 steps.}

\section{Surgery-Inspired Robotic Tasks}\label{sec:surgical_tasks}
In this section, we examine the practical application of our shape servoing framework in addressing surgical robotic tasks.

\subsection{Surgical Retraction}
\begin{figure*}[t!]
  \centering
    \includegraphics[width=1.0\textwidth]{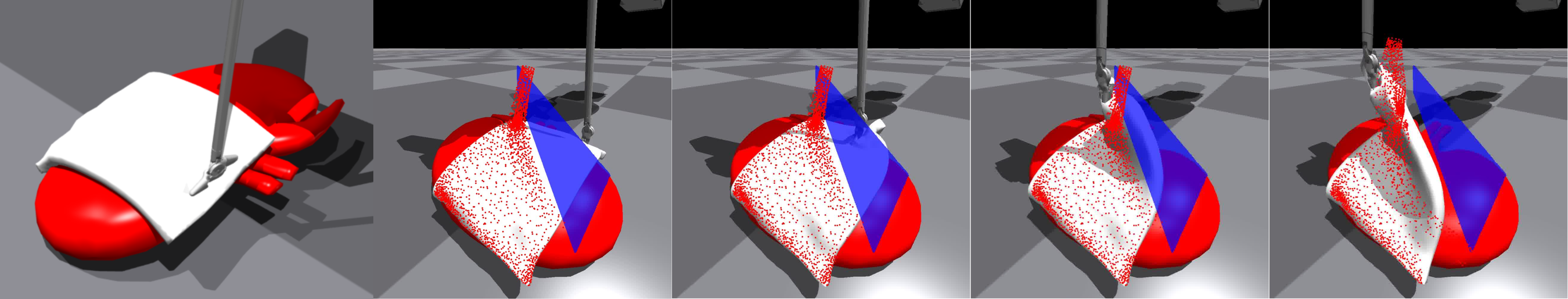} \\
    \includegraphics[width=1.0\textwidth]{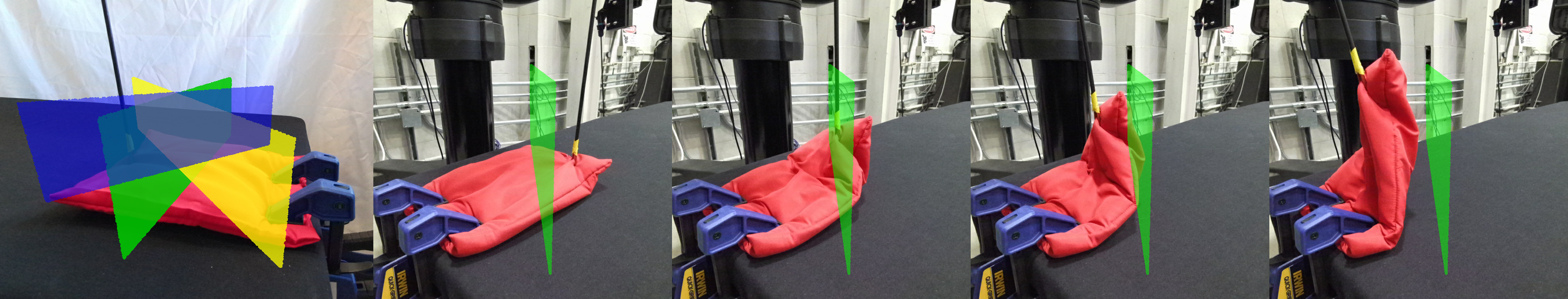}    
    \vspace{-1mm}
    \caption{\textit{Top row}: simulated retraction experiment setup (leftmost) and a sample successful retraction sequence with target plane visualized in blue. \textit{Bottom row}: visualization of target planes for physical robot retraction experiment (leftmost) and a successful sequence with target plane visualized in green.}
    \label{fig:sequence_plane}
    \vspace{-5mm}
\end{figure*}

We apply our shape servoing framework on a mock
surgical retraction task, in which a thin layer of tissue is positioned on top of a kidney, and the robot is tasked with grasping the tissue and lifting it up to expose the underlying area.
Figure~\ref{fig:sequence_plane} (top, left) shows the simulation environment composed of a kidney model with a deformable tissue layer placed over it and fixed to the kidney on one side. We train \DeformerNet{} on a box object similar in dimensions to the tissue layer, but without the kidney present. 

Instead of requiring the operator (e.g. surgeon) to provide an explicit shape for the robot to servo the tissue to, we simply ask them to define a plane which the tissue should be folded to one side of. We refer to the side of the plane where the tissue must be fully placed as the \textit{good} side, while the opposite side is named the \textit{bad} side.
An example plane can be seen in Fig.~\ref{fig:sequence_plane}.

Here we present our approach to translating the hand-specified plane into a goal point cloud interpretable by our \DeformerNet{} shape servoing algorithm. We first use the \revised {RANSAC (RANdom SAmple Consensus)~\cite{fischler1981random} to find a dominant plane in the object cloud. RANSAC does so by generating candidate optimal planes fit to a number of random subsets of points in the point cloud and evaluating how well the planes fit to the entire point cloud. We then find the minimum rotation to align this plane with the target plane}. We then apply this estimated transform to any points not lying on the correct side of the plane, merge this with the points currently satisfying the goal, and set this combined point cloud as the goal point cloud. Next, we run \DeformerNet{} with the generated heuristic goal point cloud until convergence. If the robot still does not succeed at the task after convergence, we update the goal point cloud using the following procedure. First, we shift the target plane by a small amount toward the \textit{good} side and set this as the new target plane. We then repeat the process above to obtain a new goal point cloud and run it with \DeformerNet{}. We iteratively update the heuristic goal and execute our shape servoing framework until the entire tissue layer resides in the \textit{good} side of the plane, and the task is considered successful.
However, if during the iterative target plane updates, the object is found to lie entirely on the \textit{bad} side of the plane, we terminate the heuristic goal generation process and deem the task unsuccessful.

To evaluate, we sample 100 random planes with different orientations in simulation and task the method with moving the tissue beyond the planes. 
Our approach accomplishes the task with a success rate of 95\%.

We also evaluate retraction on the physical robot.
We affix a thin layer of foam to a table and task the robot with moving the object via the laparoscopic tool beyond a target plane.
We evaluate on 3 different planes (see Fig.~\ref{fig:sequence_plane} bottom left), and for each plane conduct 5 trials.
We observe a 100\% success rate across the 15 trials.
We provide visualizations of representative retraction experiments, both in simulation and on a physical robot, in Fig.~\ref{fig:sequence_plane}.

\subsection{Connecting Tube-like Objects}

This task involves two robot arms connecting the ends of two tube-like objects together. 
Such a task is critical in a surgical procedure called anastomosis where two anatomical lumens (e.g, blood vessels, intestine portions, etc.) need to be connected together and then sutured.
We treat this task as two independent single-arm manipulation problems of two separate deformable objects. By conducting this experiment, our goal is to demonstrate an intriguing application scenario in which two instances of \DeformerNet{} can coordinate to achieve a task.

\begin{figure}[ht]
    \centering
    \includegraphics[width=0.49\linewidth]{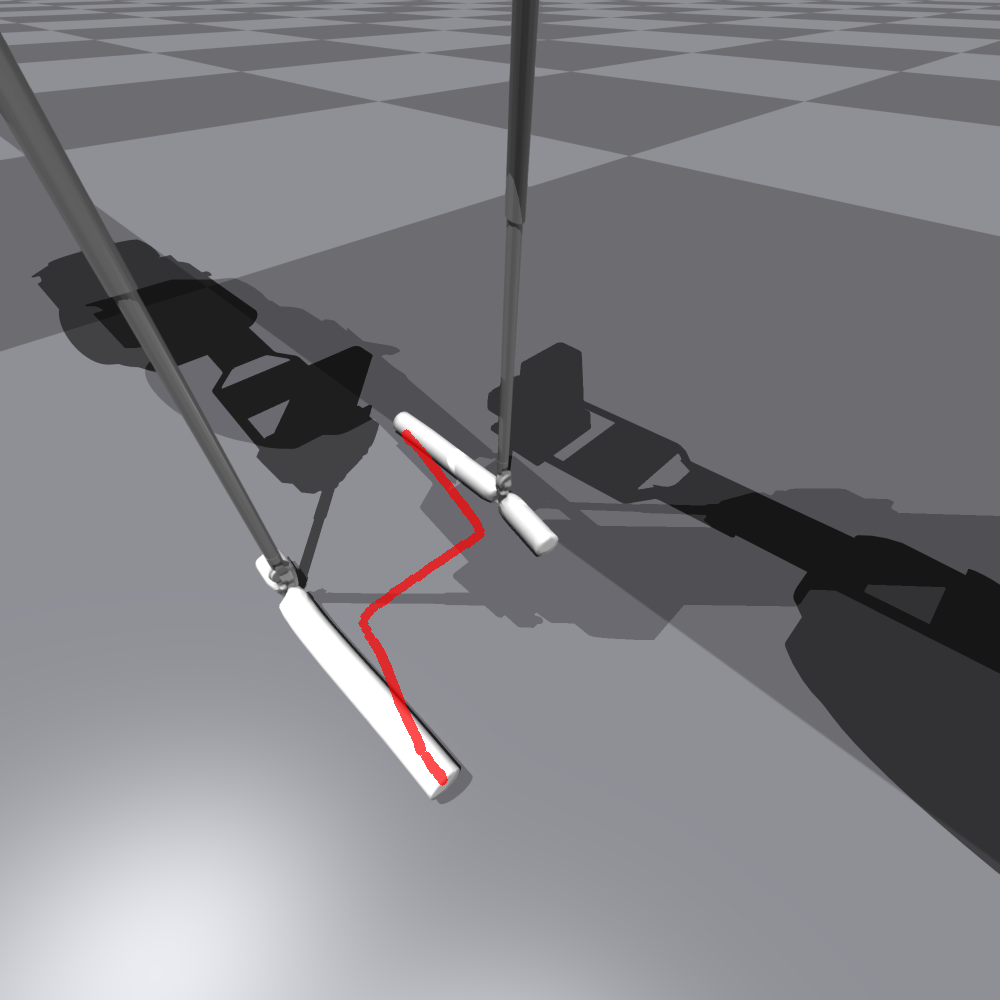}%
    \includegraphics[width=0.49\linewidth]{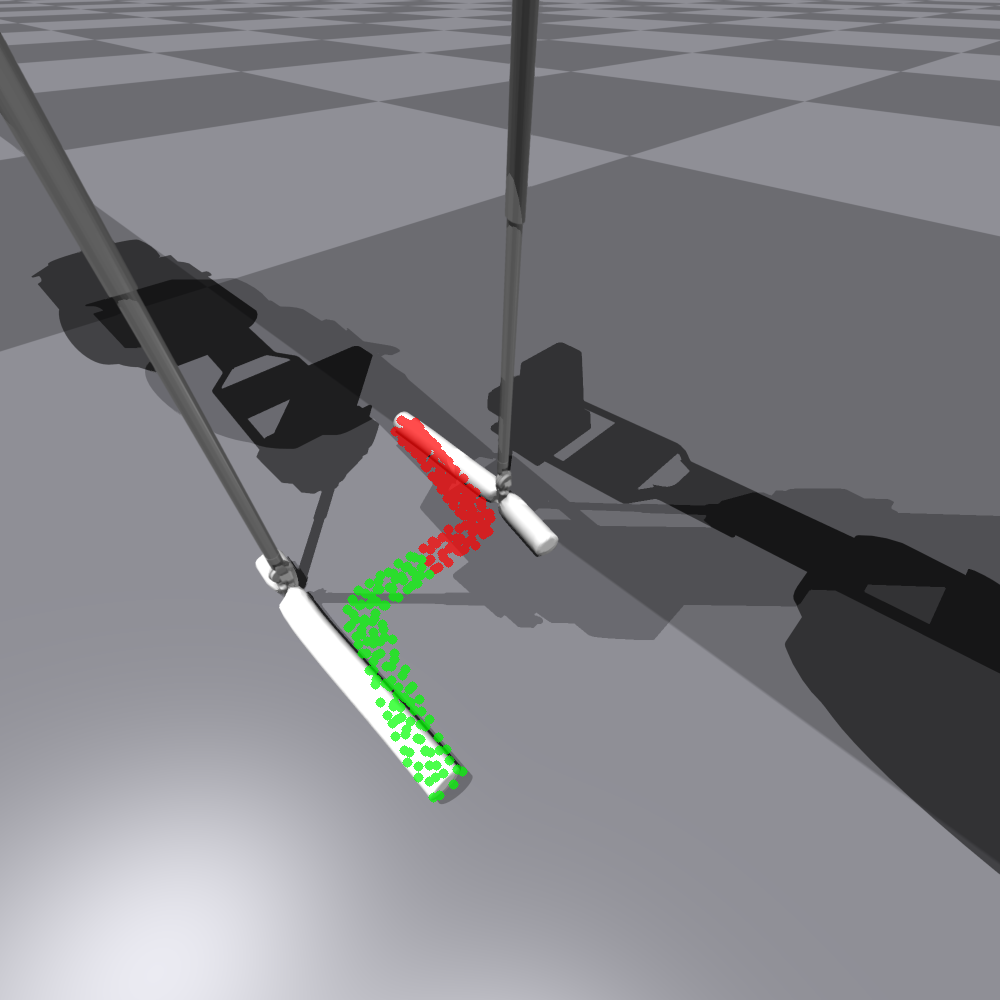} \\
    \includegraphics[width=0.49\linewidth]{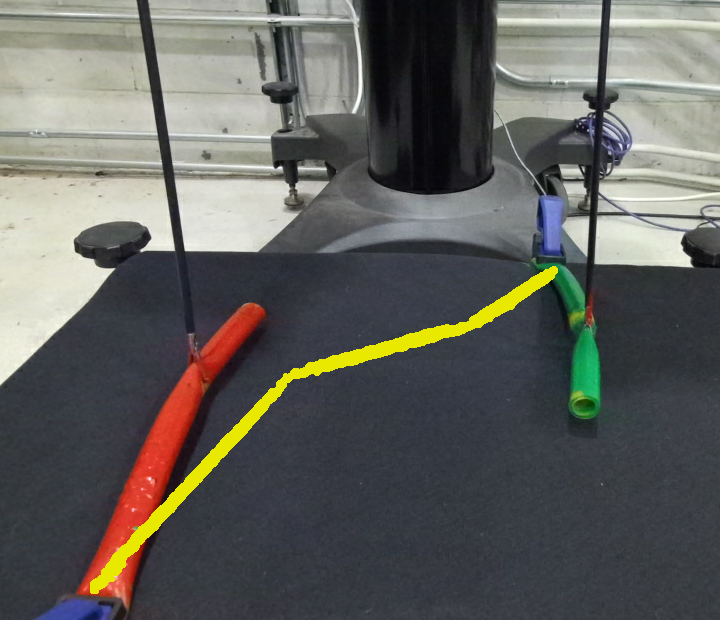}%
    \includegraphics[width=0.49\linewidth]{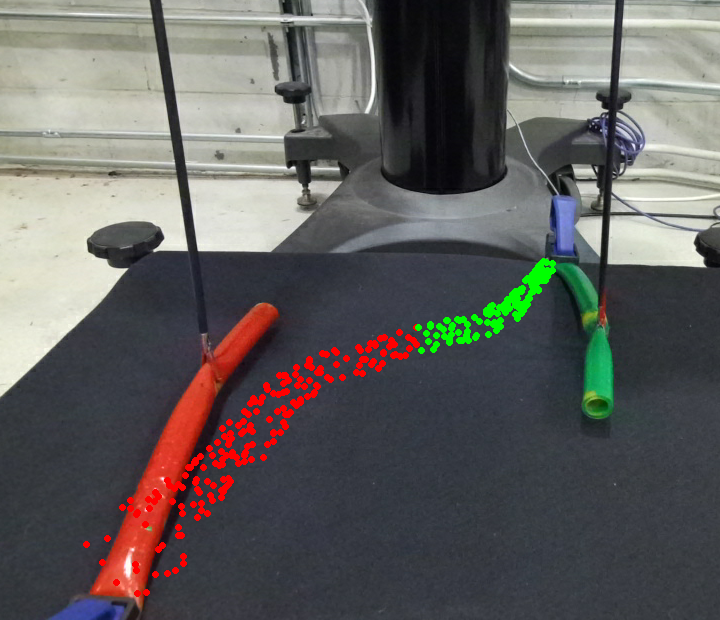}%
    \caption{Example input space curves for tube connecting task and the corresponding heuristic goal point clouds. \textit{(Top row)} In simulation. \textit{(Bottom row)} In physical robot experiment.}
    \label{fig:tube_connect_drawing}
    \vspace{-5mm}
\end{figure}

We obtain the heuristic goal point cloud from a 3D space curve given by a human user (e.g., via a 3D mouse or haptic device) which describes how they wants the two tubes to be connected. In Fig.~\ref{fig:tube_connect_drawing}, we present examples of the curves used in our experiments as well as the corresponding generated heuristic goal point clouds. The curve is utilized as the skeleton to create an imaginary goal tube that has diameters equal to those of the tubes in the environment. 
We symmetrically split this into two halves, then uniformly sample points on these two goal tubes to create two heuristic goal point clouds. These goals are then fed separately to two $\DeformerNet{}$ instances, which output two manipulation actions for each of the robot arms. 

\begin{figure*}[t!]
  \centering
    \includegraphics[width=1.0\textwidth]{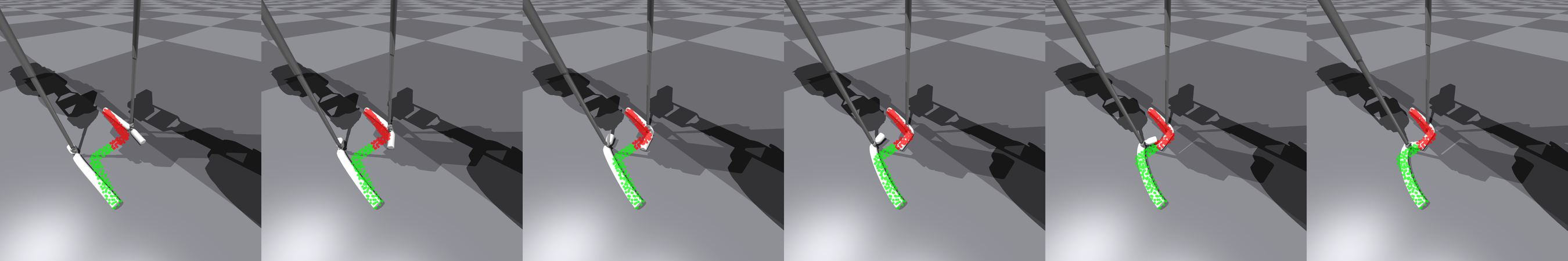} \\
    \includegraphics[width=1.0\textwidth]{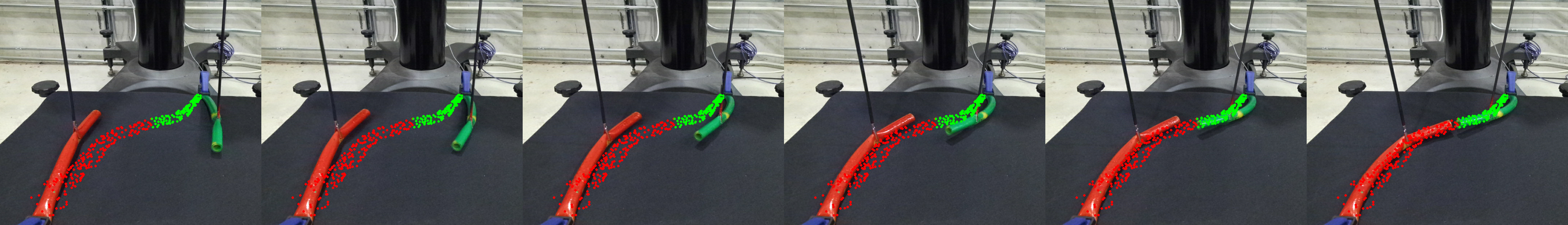}  
    \includegraphics[width=1.0\textwidth]{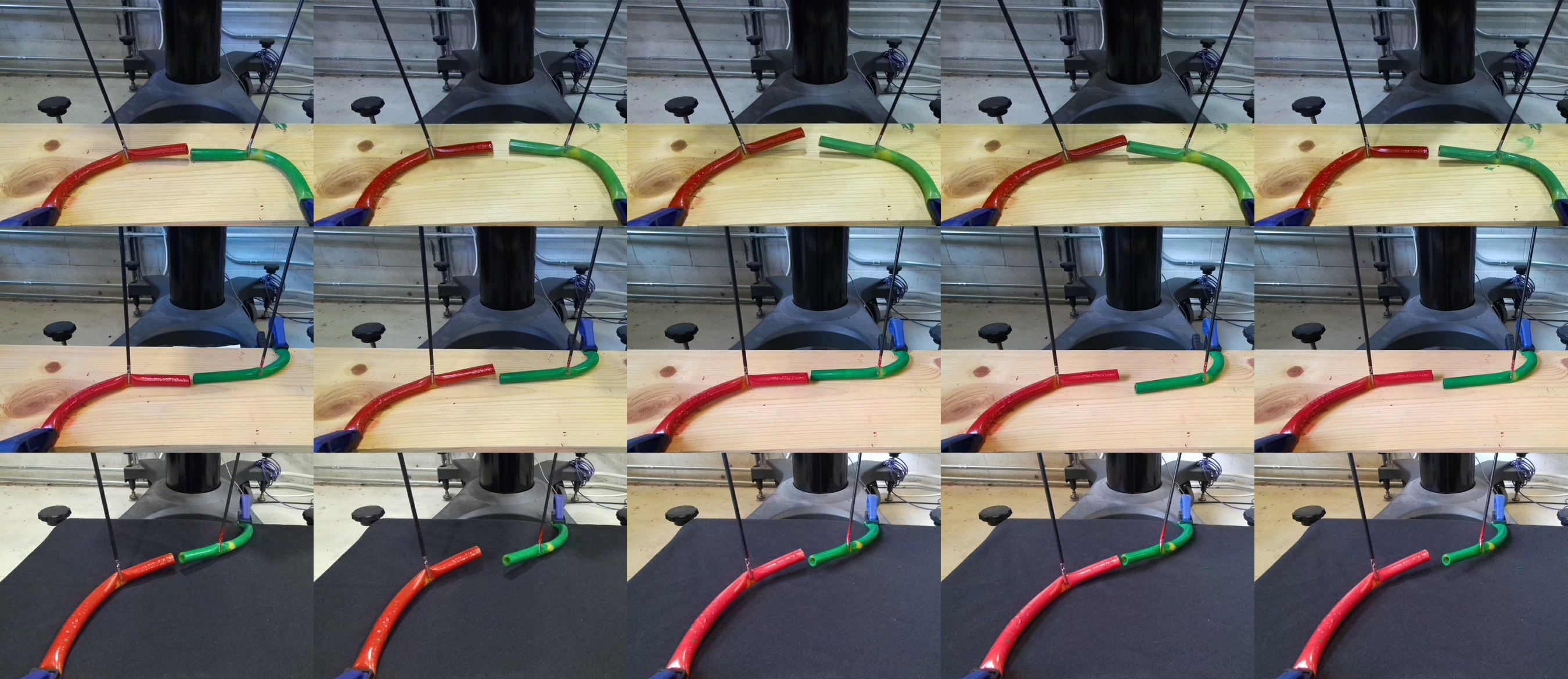} 
    \vspace{1pt}
    \caption{\textit{Top row}: A sample successful tube connecting sequence in simulation with the heuristic goal point cloud visualized. \textit{Second row}: A sample successful tube connecting sequence on the physical robot.
    \textit{Last three rows}: Final shapes of the tubes over 15 experiment runs.}
    \vspace{-6mm}
    \label{fig:sequence_sim_tube_connect}
\end{figure*}

In simulation, we design two evaluation metrics for this task. The first one is the position difference between the two ends of the tubes. We compute this metric by tracking the centers of the two ends and computing the Euclidean distance between them. The second one is the total Fr\'{e}chet distance between the final backbones of the two tubes and the input curve. These two metrics together measure how well \DeformerNet{} performs in connecting the two tube ends as well as in matching their shapes with the drawing curve. To evaluate, we run our experiment on three distinct input drawing curves, each with 100 trials where the tubes are initialized at various random shapes.
The lowest recorded tube-end position difference was 0.003 meters, the highest was 0.089 meters, the mean was 0.027 meters, and the standard deviation was 0.016 meters. The lowest recorded Fr\'{e}chet distance was 0.041 meters, the highest was 0.148 meters, the mean was 0.086 meters, and the standard deviation was 0.026 meters. We provide visualization of a representative manipulation sequence in Fig.~\ref{fig:sequence_sim_tube_connect} (first row).  

For the physical robot experiment, we affix two cylindrical tubes to a table and task the robot with connecting them via the laparoscopic tool. 
We evaluate our methods on 3 different input curves, and for each curve conduct 5 trials. 
We provide visualization of a representative manipulation sequence in the second row of Fig.~\ref{fig:sequence_sim_tube_connect}. We also show the final shapes of the tubes over 15 experiment runs in the last three rows of Fig.~\ref{fig:sequence_sim_tube_connect}. As observed from the visualized final shapes, our method does not perfectly align the two tube ends. However, the tubes qualitatively match with the heuristic goal point clouds very well, and the two tube ends almost align with each other in all cases.

\subsection{Tissue Wrapping}

This surgical task involves two robot arms coordinating to wrap a thin tissue layer around a cylindrical tube, covering as much surface of the tube as possible. 
The tissue wrapping task is inspired by surgical procedures such as aortic stent placement.
We treat this task as a bimanual manipulation of a deformable object, running a single instance of the bimanual \DeformerNet{}. 

\begin{figure*}[t!]
  \centering
    \includegraphics[width=1.0\textwidth]{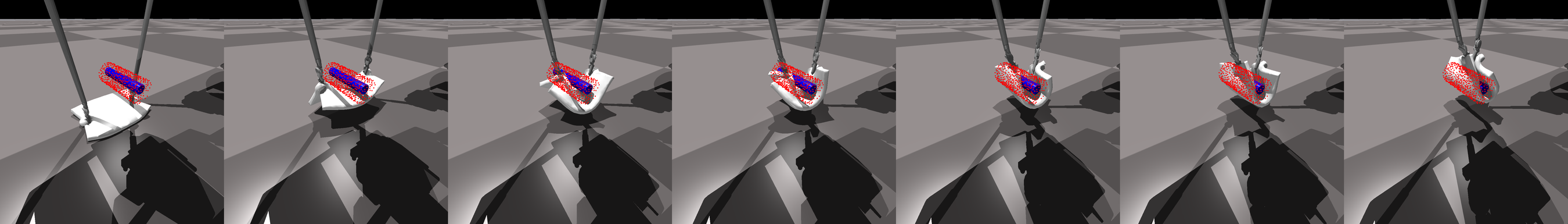} \\
    \includegraphics[width=1.0\textwidth]{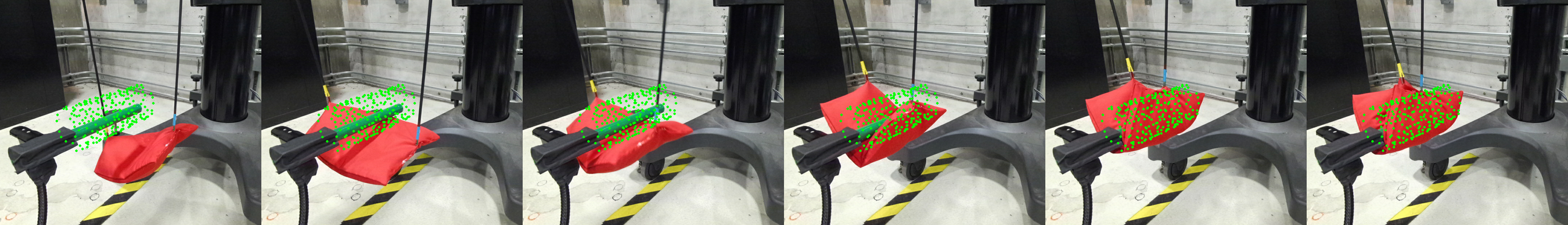}  %
    \includegraphics[width=1.0\textwidth]{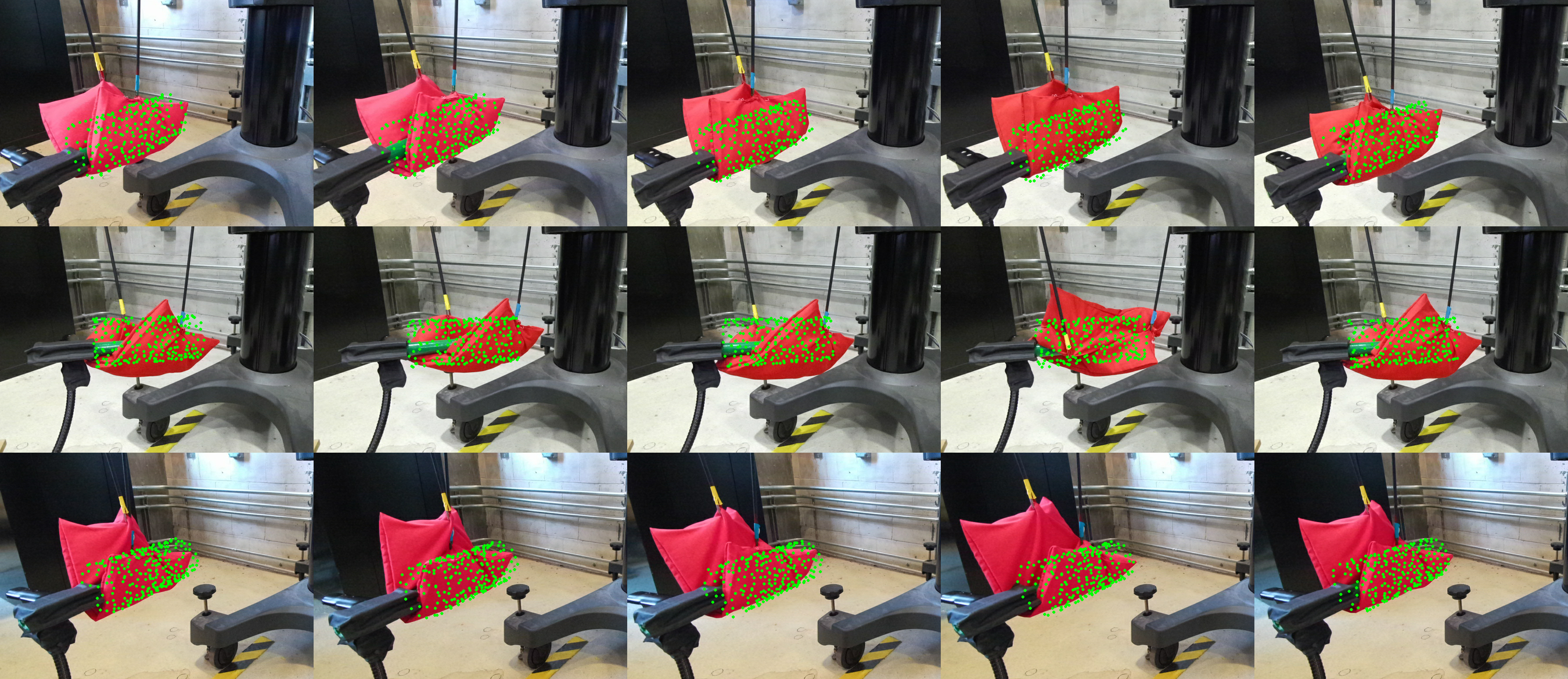} 
    \vspace{-2mm}
    \caption{\textit{Top row}: A sample successful tissue wrapping sequence in simulation, with the target tube and heuristic goal point cloud visualized in blue and red respectively. \textit{Second row}: A sample successful tissue wrapping sequence on the physical robot.
    \textit{Last three rows}: Final shapes of the box-like object over 15 experiment runs.}  
    \vspace{-6mm}
    \label{fig:sequence_sim_tissue_wrap}
\end{figure*}

We obtain the heuristic goal point cloud by first generating a goal cylinder
with pose and length same as the target tube, but with circumference equal to the length of the tissue (see first row of Fig.~\ref{fig:sequence_sim_tissue_wrap}). We then uniformly sample points on the surface of this goal cylinder and set this to be the heuristic goal point cloud. This goal intuitively encourages a final shape of the tissue such that the tissue covers the entire surface of the cylindrical tube. 
Once the heuristic goal is computed, the task is executed in two steps. First, we move the tissue such that the centroid of the tissue aligns with the centroid of the target cylindrical tube. Second, we sense the current point cloud of the tissue and apply the bimanual $\DeformerNet$ to control the shape of the tissue. 

In simulation, we design an evaluation metric that measures the percentage of the cylindrical tube surface being covered by the tissue.
We compute this metric by first evenly sampling points on the tube surface, and then projecting rays out from these points. The direction of the ray coming out from each point is set to be the same as the normal vector of that point.
We then count the percentage of rays that intersect with the wrapped tissue. 
We run our experiment on 100 distinct target tube poses.
A representative manipulation sequence is visualized in Fig.~\ref{fig:sequence_sim_tissue_wrap} (first row).
The lowest recorded coverage percentage was 79.0\%, the $25^\textrm{th}$ percentile was 91.3\%, median was 95.1\%, $75^\textrm{th}$ percentile was 98.0\%, and the highest was 100\%. The average coverage percentage was 93.80\%. This means that our method, on average, can manipulate the tissue to cover more than 90\% of the surface of the target tube.

For the physical robot experiment, we use a PVC pipe for the target cylindrical tube and a soft box-like deformable object in place of the biological tissue.
We assess the performance of our approach on 3 distinct target tube poses (see Fig.~\ref{fig:sequence_sim_tissue_wrap}), and for each pose, we conduct 5 trials where the box-like deformable object is initialized in various poses.
We provide visualization of a representative experiment in the second row of Fig.~\ref{fig:sequence_sim_tissue_wrap}. We also show the final shapes of the task over 15 experiment runs in the last three rows of Fig.~\ref{fig:sequence_sim_tissue_wrap}. 
As observed from the visualized final shapes, our method qualitatively succeeds in wrapping the object around the cylinder in all cases.

\section{Discussion and conclusions}\label{sec:conclusions}
In this paper, we present a novel learning-based approach to closed-loop 3D deformable object shape control. Through rigorous simulated and physical-robot experiments, we demonstrate that our shape servoing framework with \DeformerNet{} can effectively generalize what it learns from training to adapt to various geometries, material properties, and goal shapes. Furthermore, we show how our shape servoing approach can be applied to the surgery-inspired tasks of surgical retraction, tissue wrapping, and tube connecting, where only a simpler goal representation needs to be provided. 

However, it is important to note that despite qualitatively succeeding in most test goal shapes, some edge cases still exist where $\DeformerNet$ cannot successfully accomplish the task. We have carefully investigated the top 10\% worst performing goal shapes of each object category and arrived at the following common failure cases. First (most common), $\DeformerNet$ might converge to shapes quite close to (but not at) the goal and get stuck at these local minima, especially 
 for complex goal shapes that require a large end-effector displacement to achieve. Please refer to the two rightmost images in Fig.~\ref{fig:single_quartiles_sim}, ~\ref{fig:bimanual_quartiles_sim}, ~\ref{fig:single_quartiles_real} and~\ref{fig:bimanual_quartiles_real} for examples of this failure mode.
 Second, there are a few edge cases where self-occlusion causes the camera to only see a very small portion of the goal shape. 
 Third, there are cases where the two robot arms accidentally collide with each other. We stop the robot at collision instances, hence resulting in a poor final shape.

Our future work aims to develop a method that allows $\DeformerNet$ to iteratively fine-tune the final shapes to match the complex goals better.
We also wish to extend our manipulation approach to more surgical tasks, to the manipulation of materials that plastically deform~\cite{huang2021plasticinelab}, and to manipulating 3D deformable objects common to homes and warehouses. Finally, we aspire to move beyond our local visual servoing approach to establish a more comprehensive planning paradigm for longer-horizon tasks.

\section*{Acknowledgement}
This work was supported in part by NSF Awards \#2024778 and \#2133027.

\bibliographystyle{IEEEtran}
\bibliography{bibliography}

\end{document}